\documentclass{article}

\usepackage[preprint]{neurips_2023}
\usepackage{natbib}
\setcitestyle{numbers,square}
\usepackage[utf8]{inputenc} % allow utf-8 input
\usepackage[T1]{fontenc}    % use 8-bit T1 fonts
\usepackage{hyperref}       % hyperlinks
\usepackage{url}            % simple URL typesetting
\usepackage{booktabs}       % professional-quality tables
\usepackage{amsfonts}       % blackboard math symbols
\usepackage{nicefrac}       % compact symbols for 1/2, etc.
\usepackage{microtype}      % microtypography
\usepackage{xcolor}         % colors
\usepackage{graphicx}
\usepackage{wrapfig}
\usepackage{multirow}
\usepackage{epsfig}
\usepackage{amsmath}
\usepackage{amssymb}
\usepackage{subfigure}
\usepackage{marvosym}
\usepackage{color}
\usepackage{threeparttable}
\usepackage{algorithm}
\usepackage{algpseudocode}
\usepackage{setspace}
\usepackage{amsthm}
\usepackage[hang,flushmargin]{footmisc}

\newcommand{\update}[1]{{\textcolor{black}{#1}}}
\newcommand{\boldres}[1]{{\textbf{\textcolor{black}{#1}}}}
\newcommand{\secondres}[1]{{\underline{\textcolor{black}{#1}}}}

\title{Client: Cross-variable Linear Integrated Enhanced Transformer for Multivariate Long-Term Time Series Forecasting}

\author{%
  Jiaxin Gao\\
  Shanghai Jiaotong University\\
  Eastern Institute for Advanced Study \\
  \texttt{jiaxingao@sjtu.edu.cn} \\
  \And
  Wenbo Hu \\
  Hefei University of Technology \\
  \texttt{wenbohu@hfut.edu.cn} \\
  \AND
  Yuntian Chen \\
  Eastern Institute for Advanced Study \\
  \texttt{ychen@eias.ac.cn} \\
}

\begin{document}

\maketitle

\begin{abstract}
Long-term time series forecasting (LTSF) is a crucial aspect of modern society, playing a pivotal role in facilitating long-term planning and developing early warning systems. While many Transformer-based models have recently been introduced for LTSF, a doubt have been raised regarding the effectiveness of attention modules in capturing cross-time dependencies. In this study, we design a mask-series experiment to validate this assumption and subsequently propose the "Cross-variable Linear Integrated ENhanced Transformer for Multivariate Long-Term Time Series Forecasting" (Client), an advanced model that outperforms both traditional Transformer-based models and linear models. Client employs linear modules to learn trend information and attention modules to capture cross-variable dependencies. Meanwhile, it simplifies the embedding and position encoding layers and replaces the decoder module with a projection layer. Essentially, Client incorporates non-linearity and cross-variable dependencies, which sets it apart from conventional linear models and Transformer-based models. Extensive experiments with nine real-world datasets have confirmed the SOTA performance of Client with the least computation time and memory consumption compared with the previous Transformer-based models. Our code is available at \url{https://github.com/daxin007/Client}.
\end{abstract}

\section{Introduction}
Long-term time series forecasting (LTSF) is a critical element in today's data-driven world, with applications in a variety of fields, such as energy management \cite{gao2023adaptive}, stock price prediction \cite{lopez2023can}, traffic flow prediction \cite{fang2022attention}, and weather forecasting  \cite{meenal2022weather}, to name a few. Recent advancements in deep-learning models, notably Transformer-based models, have shown significant progress in this field and offer immense potential for addressing the multifaceted challenges associated with long-term forecasting~\cite{wen2022transformers}.

There is a question about the effectiveness of the Transformer model for LTSF \cite{Zeng2022AreTE}. The previous Transformer models took advantage of self-attention to capture the long-term dependencies of different time steps for LTSF. For instance, Informer \cite{haoyietal-informer-2021} leverages KL-divergence based ProbSparse attention to expand the Transformer model, and Autoformer \cite{wu2021autoformer} replaces the usual self-attention mechanism with an Auto-Correlation module, which achieves $O(L\log L)$ complexity. Nevertheless, DLinear~\cite{Zeng2022AreTE} argues that most previous Transformer models based on cross-time attention fail to adequately learn cross-time dependencies and demonstrates that a simple linear model outperforms most existing Transformer models. 
 
To better understand this issue, we introduce a comprehensive mask-series experiment and randomly set a fraction of the historical series to $0$. 
Models that are more sensitive to cross-time dependency suffer greater performance degradation in the absence of historical data. Therefore, the performance degradation is an indicator for model's representation capability of cross-time patterns. As can be seen in Figure~\ref{fig:mask_pred}, the cross-time attention based Transformer models' performance does not significantly decrease as the mask scale increases. Among these models, the prediction performance of Informer~\cite{haoyietal-informer-2021} remains unchanged even when $80\%$ of historical values are randomly set to 0, which indicates that it is not sensitive to cross-time dependency.This result suggests that the structure of cross-time attention may be inadequate for learning trend information in time series effectively, which is coincident with the results in DLinear~\cite{Zeng2022AreTE}. We also observe that in some multivariate time series, different variables exhibit related patterns over time. This suggests that it may be feasible to use attention to learn dependencies between variables rather than across time steps. Hence, in this paper, we dispel the aforementioned doubt by replacing the conventional cross-time Transformer with the cross-variable Transformer.

Though the proposed Transformer is good at modeling non-linearity and capturing cross-variable dependencies, they may fail in extracting series trends, at which the linear models excel~\cite{Zeng2022AreTE}.
To combine the best of the both worlds, we propose "Cross-variable Linear Integrated Enhanced Transformer for Multivariate Long-Term Time Series Forecasting", Client for short, which combines the trend-extraction ability of linear models with the expressive capacity of enhanced Transformer models for forecasting the complex details in time series. 

In summary, the contribution of this study is threefold:
\begin{itemize}
    \item This study designs a comprehensive mask-series experiment to verify the effectiveness of conventional Transformer models in capturing cross-time dependencies and demonstrate that the models are ineffective in capturing information from time series adequately. 
    \item This study proposes Client, which combines a linear model to learn the linear trends and a cross-variable Transformer model to learn non-linear information and dependencies between variables, and discards the cross-time attention of Transformer. The enhanced Transformer model also simplifies the embedding and position encoding layers and replaces the decoder module with a projection layer, which significantly improves the forecasting performance. Furthermore, Client proves that cross-variable attention in Transformer is more important than cross-time attention for LTSF.
    \item Client achieves SOTA performances on multiple real-world datasets with the least training time and memory consumption compared with the conventional Transformer models and the previous SOTA model, TimesNet. Moreover, Client demonstrates better forecasting results when the look-back (historical) window size grows, which proves that Client not only has higher accuracy, but also has stronger ability to learn long-term dependency in time series. In contrast, the previous cross-time Transformer models and TimesNet would not perform better with an increase of the look-back window size.
\end{itemize}

\section{Related Work}
The field of multivariate time series forecasting has seen the development of several models, which can be generally categorized into Non-Transformer and Transformer-based models.
\subsection{Non-Transformer models}
Traditional statistical models such as ARMA and ARIMA \cite{box2015time} have been widely used in time series forecasting, which assumes linear relationships between past and present observations to capture trends in the series. However, with the amount of available big data and the advancement of deep learning techniques, there has been a recent shift towards using deep learning models for time series forecasting, and deep learning models often exhibit better performance than traditional statistical models thanks to their more effective expressive ability and better utilization of data.

Some of the most popular deep learning models used for time series forecasting include Multilayer Perceptron (MLP) \cite{1991Multilayer}, Recurrent Neural Networks (RNNs) \cite{Hochreiter1997LongSM}, Convolutional Neural Networks (CNNs) \cite{1998Gradient}, and Graph Neural Networks (GNNs) \cite{2009The}. DeepAR combines RNN and autoregressive methods to model the probabilistic features of the series~\cite{Flunkert2017DeepARPF,cui2020calibrated}. Some temporal convolution networks (TCN) based works \cite{Oord2016WaveNetAG,Sen2019ThinkGA,zhang2020dynamic,liu2022scinet} attempt to model the temporal causality with the causal convolution. LSTnet \cite{2018Modeling} exploits CNN and RNN to extract short-term local dependency and long-term global correlation. MTGNN \cite{wu2020connecting} uses TCN and GCN to capture cross-time and cross-variable dependencies. Dlinear \cite{Zeng2022AreTE} is a classical work which uses a linear model to achieve good results in LTSF. LightTS \cite{Zhang2022LessIM} is a lightweight MLP-based model that uses interval sampling and continuous sampling to extract features from time series. TimesNet \cite{wu2023timesnet} transforms the 1D time series into a set of 2D tensors based on multiple periods and uses a CNN-based model to extract features from 2D tensors, it achieves state-of-the-art in LTSF and four other mainstream time series tasks.
\subsection{Transformer models}
Transformer \cite{2017attention} is widely recognized as a highly successful model architecture, delivering impressive results across various fields, including NLP, speech recognition, and computer vision \cite{devlin2018bert} \cite{brown2020language} \cite{gulati2020conformer} \cite{liu2021swin}.Several Transformer variants have been introduced to improve the self-attention mechanism's design for time series forecasting. These models aim to keep time and space complexity at a minimum while improving performance. One such variant is LogTrans \cite{2019Enhancing}, which uses convolutional self-attention layers with a LogSparse design to capture local information while lowering space complexity. Informer \cite{haoyietal-informer-2021} and Autoformer \cite{wu2021autoformer} reduce complexity to $O(L\log L)$ through replacing the usual self-attention mechanism. Through the utilization of Fourier or wavelet transforms and randomly selecting frequency bases, FEDformer \cite{zhou2022fedformer} reduces complexity to $O(L)$ by applying frequency-domain self-attention. Another alternative, Pyraformer\cite{liu2021pyraformer}, incorporates a pyramidal attention module with inter-scale and intra-scale connections, which also produces a linear complexity. ETSformer \cite{woo2022etsformer} replaces the self-attention mechanism with exponential smoothing attention and frequency attention to improve the model's accuracy and efficiency. Non-stationary Transformer \cite{Liu2022NonstationaryTR} incorporates stationarization modules and De-stationary Attention into vanilla Transformer to make the model's prediction more stationary while avoiding over-stationarization. 
These Transformer models all use cross-time attention to learn long-term dependency, and it is doubted by DLinear~\cite{Zeng2022AreTE} that cross-time attention can not learn time treads effectively. This paper proposes the mask-series experiment and validates the ineffectiveness of the cross-time Transformer models. Crossformer \cite{zhang2023crossformer} utilizes a Two-Stage Attention (TSA) layer to capture both the cross-time dependencies and the dependencies of series segments across different dimensions. However, Crossformer has a similar or worse performance on most benchmarking datasets compared with DLinear. The detailed analysis of CrossFormer can be found in the Appendix \ref{app:crossformer}.

\section{Client}

The long-term time series forecasting (LTSF) is to predict the long-term future values of time series data. 
Given a historical series ${H=\{X^{t}_{1},...,X^{t}_{C}\}^{L}_{t=1}}$, where $L$ denotes the look-back window size of $C$ variables, and ${X^{t}_{i}}$ represents the value of the $i^{th}$ variable at the $t^{th}$ time step, the task is to forecast the future values of the same series, denoted as ${F=\{X^{t}_{1},...,X^{t}_{C}\}^{L+T}_{t=L+1}}$, where $T$ represents the number of time steps to forecast. The basic idea of Client is to replace the cross-time attention with the cross-variable attention and integrate a linear module in the model, to better utilize the variable dependencies and trend information, respectively. In this section, we detail the Client model components.
% \vspace{-10pt}
\subsection{Cross-variable Transformer}

\begin{figure}[!htbp]
\vspace{-10pt}
  \includegraphics[scale=0.3]{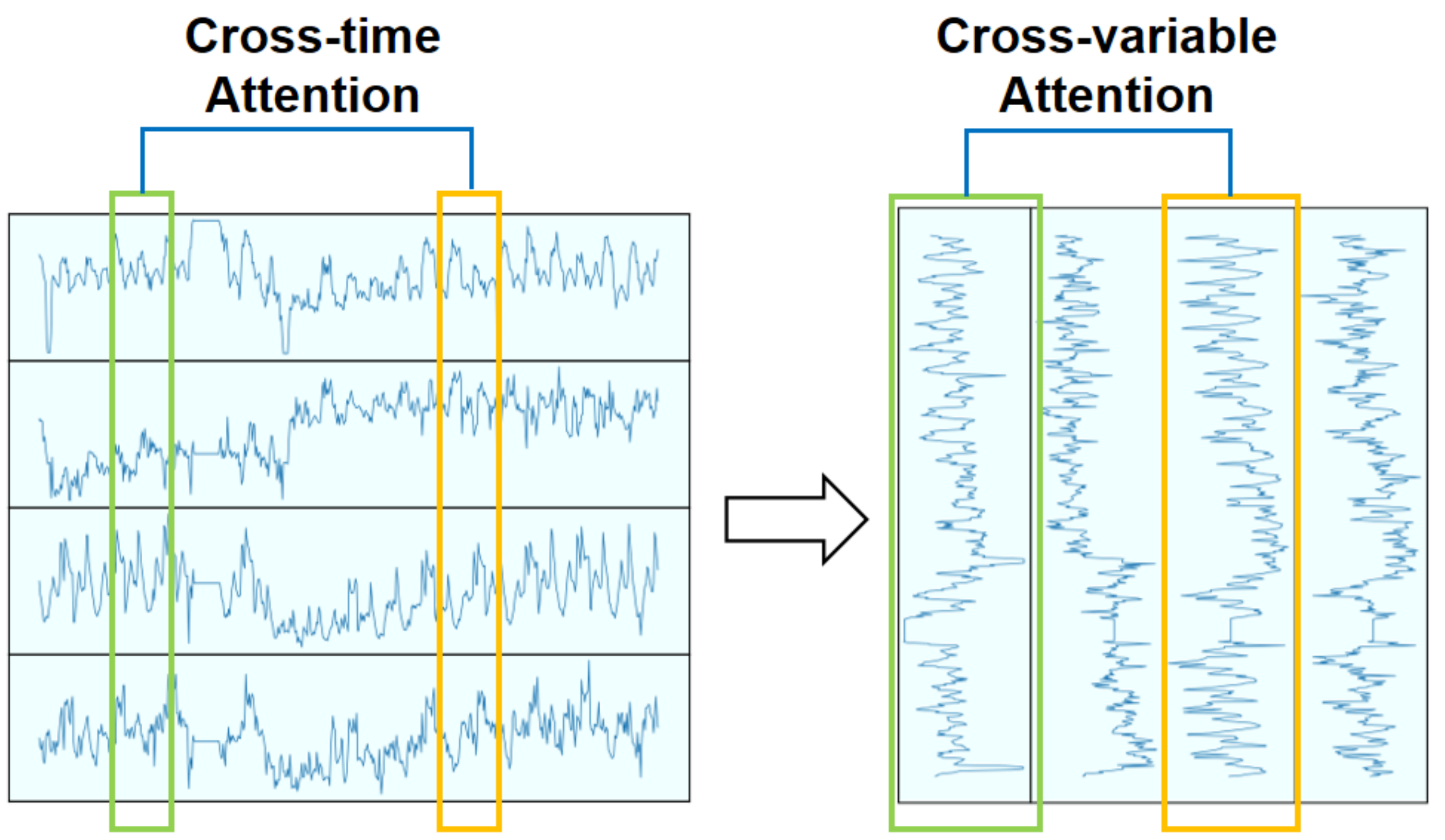}
  \centering
  \caption{{Cross-variable attention.}} 
  \label{fig:new_att}
  % \vspace{-45pt}
\end{figure}

The cross-variable Transformer module is used to learn variable dependencies in place of time dependencies, as shown in Figure \ref{fig:new_att}.

The encoder block consists of a multi-head attention (MHA) component and a feed-forward network (FFN) component. The input look-back time series ${H}$ is a 2D Tensor with the shape of ${L \times C}$, The input series needs to be flipped first. The cross-variable attention is the key part of MHA, which is defined as Eq. \ref{equ:attn}:
\begin{equation}
\label{equ:attn}
    \operatorname{Attention}(\mathbf{Q}, \mathbf{K}, \mathbf{V} )=\operatorname{softmax}\left(\frac{\mathbf{Q} \mathbf{K}^\top} {\sqrt{C}}\right) \mathbf{V}  
\end{equation}
where $\mathbf{Q}$ is queries, $\mathbf{K}$ is keys, and $\mathbf{V}$ is values. $\mathbf{Q},\mathbf{K}$ and $\mathbf{V}$ are generally obtained by applying some transformations to the original input ${H}$, and ${C}$ is number of variables. The input series is put into the Encoder blocks directly, without the embedding layer because the extra embedding layer compromises temporal information and results in inferior performance, as shown in \ref{para:embed}. Besides, we remove the position encoding layer in the Transformer since there are no temporal ordering among different variables.

After extracting features from Encoder blocks, the input series is put into a projection layer and flipped to get the prediction of the cross-variable Transformer, without passing a decoder block, as we find that that including a decoder leads to decreased performance. The process of projection is defined as:

\begin{equation}
\label{equ:proj}
\mathbf{F}_{\text{trans}}=\texttt{Proj}(\mathbf{X}_{\text{enc}})\texttt{.Permute}(1,0)
\end{equation}
where $\mathbf{X}_{\text{enc}}$ is the output of Encoder blocks, and $\mathbf{F}_{\text{trans}}$ is the Transformer's prediction. 
The cross-variable Transformer's prediction often contains the details of the time series.

\subsection{Linear Integration and ReVIN Modules}

The integrated linear module is used to learn trend information from the time series, and it is channel-independent. The input look-back time series ${H}$ is flipped and put into linear module to get the linear's prediction, as defined:
\begin{equation}
\label{equ:linear}
\mathbf{F}_{\text{lin}}=\texttt{Linear}(\mathbf{H}\texttt{.Permute}(1,0))\texttt{.Permute}(1,0)
\end{equation}
The cross-variable Transformer's prediction and the linear's prediction are combined with learnable weights $\mathbf{w}_\text{lin}$ to get the final prediction, as described in:
\begin{equation}
\label{equ:final}
\mathbf{F} = \mathbf{F}_{\text{trans}} + \mathbf{w}_{\text{lin}} \times \mathbf{F}_{\text{lin}}
\end{equation}
To address the issue of distribution shift, a reversible instance normalization (RevIN) \cite{kim2022reversible} module is adopted in the model, which is symmetrically structured to remove and restore the statistical information of a time-series instance and promote the model's stability during forecasting.
\subsection{Overall Client Architecture}
\begin{figure}[!htbp]
\includegraphics[width=\columnwidth]{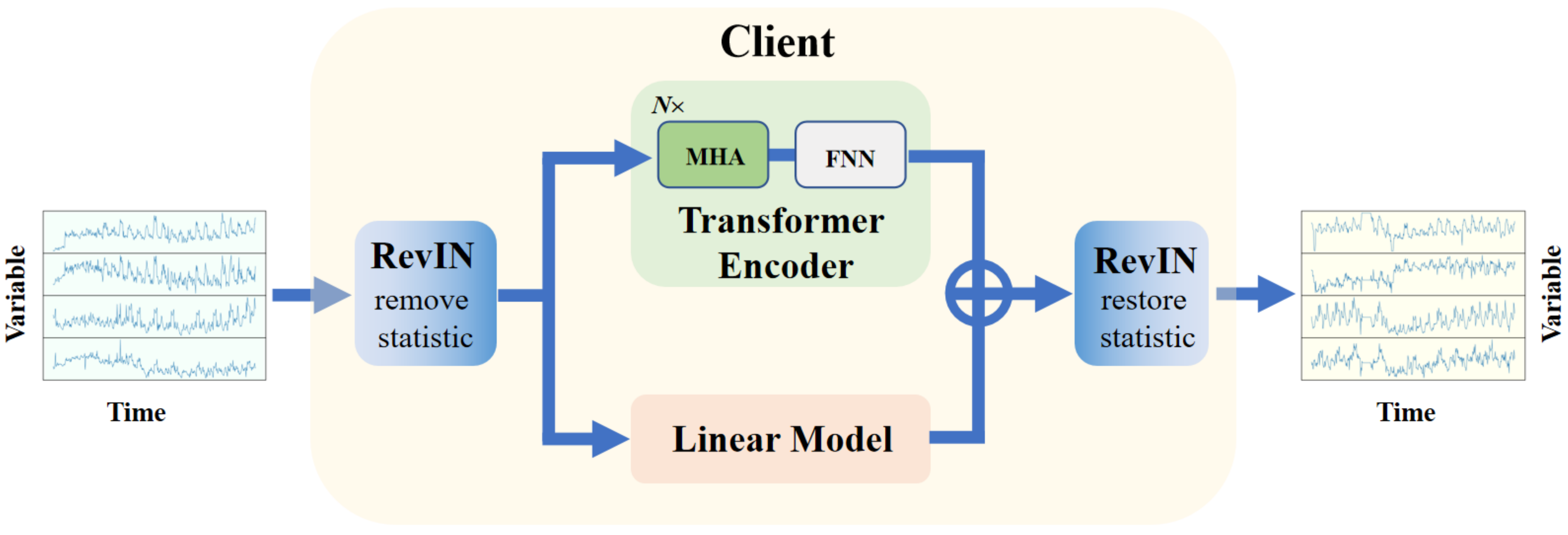}
  \centering
  \vspace{-10pt}
  \caption{Client architecture. The RevIN module is used to address the issue of distribution shift of series. The linear model is used to capture trend information, while the enhanced Transformer model is used to capture nonlinear and cross-variable dependencies.}
  \label{fig:Client}
  \vspace{-10pt}
\end{figure}

The Client model uses the linear module to capture the trend information and the enhanced Transformer module to capture nonlinear information and cross-variable dependencies. Figure \ref{fig:Client} shows the architecture of the model Client. The input series is firstly smoothed with the RevIN module. Then the smoothed series is put into the cross-variable Transformer module and the linear module respectively. The final prediction is the combination of these two module's predictions. The full algorithm of Client is presented in \ref{algo:ts}.

\begin{algorithm*}[!htbp]
  \setstretch{1.5}
  \caption{Client.}\label{algo:ts}
  \begin{algorithmic}[1]
  \Require  
  Input look-back time series: $\mathbf{x}\in\mathbb{R}^{L\times C}$; Input Length $L$; Variables number $C$; Predict length $O$; Encoder layers number $N$; Learnable coefficients of the linear model $\mathbf{w}_{\text{lin}}$.
     \State $\mathbf{x}^{\prime} = \texttt{RevIN}(\mathbf{x},encode)$
    \Comment{$\mathbf{x}^{\prime}\in\mathbb{R}^{L\times C}$}
    \State $\mathbf{x}^{\prime} = \mathbf{x}^{\prime}\texttt{.Permute}(1,0)$
    \Comment{$\mathbf{x}^{\prime}\in\mathbb{R}^{C\times L}$}
    \State $\mathbf{x}^{0\prime}_{\text{enc}}=\mathbf{x}^{\prime}$ 
    \Comment{$\mathbf{x}^{0\prime}_{\text{enc}}\in\mathbb{R}^{C\times L}$}
    \State \Comment{Cross-Variable Transformer Encoder}
    \State $\textbf{for}\ l\ \textbf{in}\ \{1,\cdots,N\}\textbf{:}$
    
    \State $\textbf{\textcolor{white}{for}}\ \mathbf{x}^{l-1\prime}_{\text{enc}} = \texttt{LayerNorm}\big(\mathbf{x}^{l-1\prime}_{\text{enc}} + \texttt{MHA}(\mathbf{x}_{\text{enc}}^{l-1\prime})\big)$
    \Comment{$\mathbf{x}^{l-1\prime}_{\text{enc}}\in\mathbb{R}^{C\times L}$}
    \State $\textbf{\textcolor{white}{for}}\ \mathbf{x}^{l\prime}_{\text{enc}} = \texttt{LayerNorm}\big(\mathbf{x}^{l-1\prime}_{\text{enc}} + \texttt{FFN}(\mathbf{x}_{\text{enc}}^{l-1\prime})\big)$
    \Comment{$\mathbf{x}^{l\prime}_{\text{enc}}\in\mathbb{R}^{C\times L}$}
    \State $\textbf{End for}$
    \State $\mathbf{x}_{\text{dec}}=\texttt{Proj}(\mathbf{x}^{l\prime}_{\text{enc}})\texttt{.Permute}(1,0)$  
    \Comment{$\mathbf{x}_{\text{dec}}\in\mathbb{R}^{O\times C}$}
    \State
    \Comment{Linear Model}
    \State $\mathbf{x}_{\text{lin}}=\texttt{Linear}(\mathbf{x}^{l\prime})\texttt{.Permute}(1,0)$  
    \Comment{$\mathbf{x}_{\text{lin}}\in\mathbb{R}^{O\times C}$}
    \State $\mathbf{y}^\prime = \mathbf{x}_{\text{dec}} + \mathbf{w}_{\text{lin}} \times \mathbf{x}_{\text{lin}}$
    \Comment{$\mathbf{y}^\prime\in\mathbb{R}^{O\times C}$}
    
    \State $\mathbf{y}^{\prime} = \texttt{RevIN}(\mathbf{y},decode)$
    \Comment{$\mathbf{y}\in\mathbb{R}^{O\times C}$}
    \State $\textbf{Return}\ \mathbf{y}$ \Comment{Return the prediction results}
  \end{algorithmic} 
\end{algorithm*} 

\section{Experiments}
\subsection{Data and experiment setting}
The performance of Client is evaluated using nine popular datasets from various fields, including Electricity \cite{ecldata}, Traffic \cite{trafficdata}, Weather \cite{weatherdata}, four ETT datasets \cite{haoyietal-informer-2021}, Exchange-Rate \cite{2018Modeling}, and ILI \cite{ilidata}. The detatiled descriptions and statistics of the datasets can be found in Appendix \ref{app:describe}.

For most datasets, the look-back window size is 96, except for ILI where it is 36 due to the amount of data. Four different prediction lengths are adopted for evaluation, which include 24, 36, 48, and 60 for ILI, and 96, 192, 336, and 720 for all other datasets. We follow the evaluation procedure utilized in a prior study \cite{haoyietal-informer-2021}, computing the mean squared error (MSE) and mean absolute error (MAE) on z-score normalized data to enable the measurement of different variables on a consistent scale.

We use three types of popular models for LTSF as baselines. 1) Transformer-based models: ETSformer \cite{woo2022etsformer}, FEDformer \cite{zhou2022fedformer}, Non-stationary Transformer \cite{Liu2022NonstationaryTR}, Autoformer \cite{wu2021autoformer}, Pyraformer \cite{liu2021pyraformer}, Informer \cite{haoyietal-informer-2021}, and LogTrans \cite{2019Enhancing}; 2) MLP(linear)-based models: LightTS \cite{Zhang2022LessIM}, and DLinear \cite{Zeng2022AreTE}; 3) CNN-based model: TimesNet \cite{wu2023timesnet}, which is also the previous SOTA model for LTSF.

In the experiments, we set the number of encoder layers to 2 and the hidden state dimension between 16 to 512. We use the ADAM optimizer \cite{DBLP:journals/corr/KingmaB14} with a learning rate between 1e-4 to 1e-2. The batch size is 32, and the training epoch is set to 10, with an early stop function of patience 3. The initial weight for the linear model $\mathbf{w}_{\text{lin}}$ is between 0.5 and 1. 
The Client model is insensitive to the hyperparameters in a large range.
All experiments can be run on a single GPU NVIDIA GeForce RTX 3090 or 4090 with a memory of 24GB.   

\begin{figure}[!htbp]
% \vspace{-15pt}
  \includegraphics[scale=0.32]{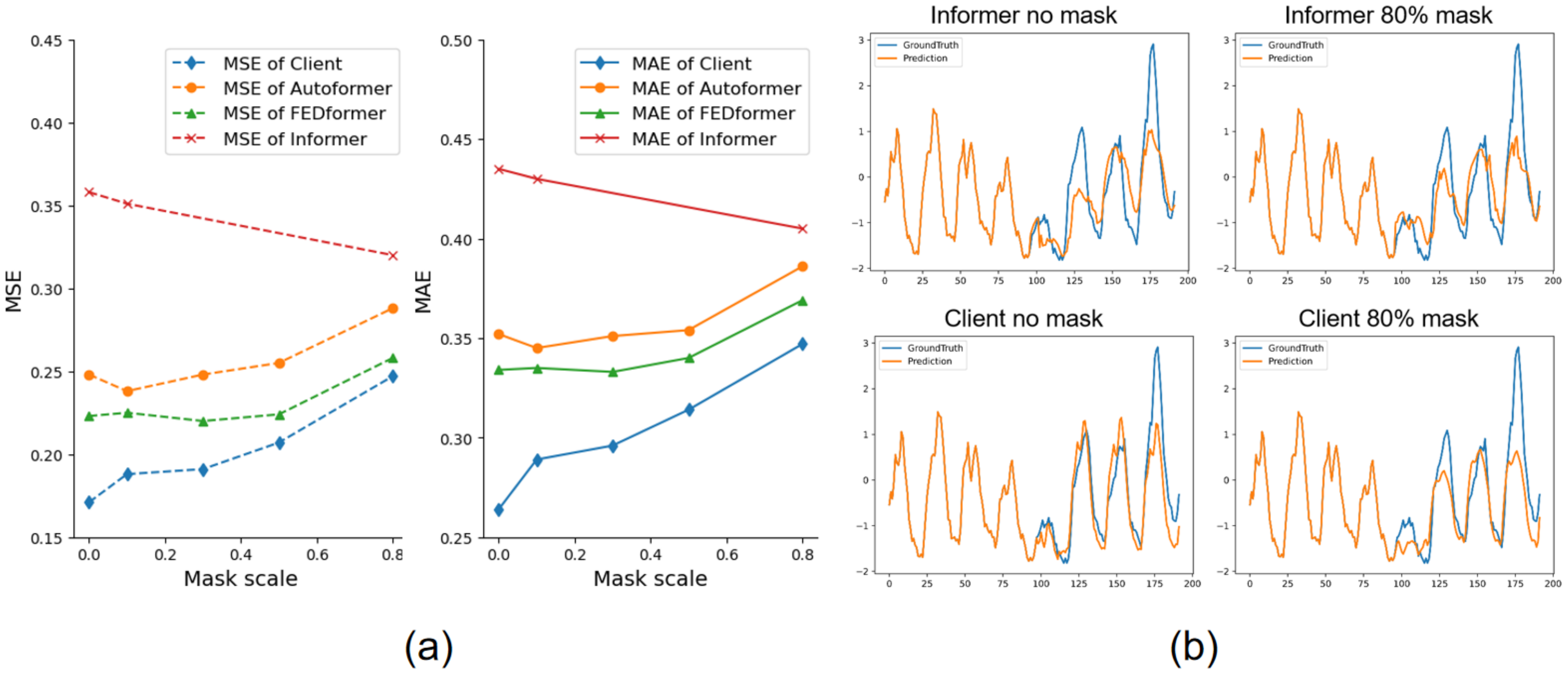}
  \centering
  \caption{(a) Masked series experiments for Transformer-based models. All the results are averaged from 4 different prediction lengths of $\{96,192,336,720\}$. With the increases in the mask scale of the input, there are clear increases in both the MSE and MAE of the Client's predictions, but there are no significant changes for other Transformer-based model's predictions. (b) Prediction showcases of Client and Informer without mask and with 80\% mask. The figure indicates that the prediction performance of Informer is not very different without mask and with 80\% mask, while Client's prediction is significantly better without mask than with 80\% mask in detail.} 
  \label{fig:mask_pred}
  \vspace{-10pt}
\end{figure}

\subsection{Mask Series Experiment}
\label{sec:mask-series}
We introduce the mask series experiment which randomly sets a fraction of the history series to 0 and the performance degradation is proportional to the representation capability of the temporal information.
We show the MSE and MAE results of Client and other Transformer-based models on the Electricity dataset in Figure \ref{fig:mask_pred}, and the results on other datasets are similar. As can be seen, the other Transformer models with cross-time attention don't have a clear performance degradation even when $50\%$ of the history series is masked.
In contrast, as the mask scale increases, there are clear increases in both the MSE and MAE of Client's predictions. Although the MSE and MAE of Client grow fastest, they are still lower than the MSE and MAE of other Transformer-based models under different mask scales.

\subsection{Main Forecasting Results}
The Long-term series forecasting results are presented in Table \ref{tab:long_term_forecasting_results}. The baseline models' results are adapted from TimesNet \cite{wu2023timesnet}. Client achieves the best performance on almost all datasets. It obtains significant improvements on several datasets with a higher number of variables (10.9\% improvement in MSE in Electricity, 23.8\% improvement in Traffic, and 3.9\% improvement in Weather), as cross-variable dependencies learned by Client are more effective with larger numbers of variables. 
\begin{table}[!htbp]
  \caption{Results of LTSF. The look-back window sizes used are 36 for ILI and 96 for the other datasets. All results are averaged across four prediction lengths: $\{24,36,48,60\}$ for ILI and $\{96,192,336,720\}$ for the other datasets. The best result is indicated in bold font, while the second-best result is underlined. Refer to Table \ref{tab:com_forecasting_results} in Appendix \ref{app:complete} for the complete results.}\label{tab:long_term_forecasting_results}
  \vskip 0.05in
  \centering
  \begin{threeparttable}
  \begin{small}
  \renewcommand{\multirowsetup}{\centering}
  \setlength{\tabcolsep}{0.78pt}
  \begin{tabular}{c|cc|cc|cc|cc|cc|cc|cc|cc|cc|cc|cc}
    \toprule
    \multicolumn{1}{c}{\multirow{2}{*}{Models}} & 
    \multicolumn{2}{c}{\rotatebox{0}{\scalebox{0.76}{\textbf{Client}}}} &
    \multicolumn{2}{c}{\rotatebox{0}{\scalebox{0.76}{TimesNet}}} &
    \multicolumn{2}{c}{\rotatebox{0}{\scalebox{0.76}{ETSformer}}} &
    \multicolumn{2}{c}{\rotatebox{0}{\scalebox{0.76}{LightTS}}} &
    \multicolumn{2}{c}{\rotatebox{0}{\scalebox{0.76}{DLinear}}} &
    \multicolumn{2}{c}{\rotatebox{0}{\scalebox{0.76}{FEDformer}}} & \multicolumn{2}{c}{\rotatebox{0}{\scalebox{0.76}{Stationary}}} & \multicolumn{2}{c}{\rotatebox{0}{\scalebox{0.76}{Autoformer}}} & \multicolumn{2}{c}{\rotatebox{0}{\scalebox{0.76}{Pyraformer}}} &  \multicolumn{2}{c}{\rotatebox{0}{\scalebox{0.76}{Informer}}} & \multicolumn{2}{c}{\rotatebox{0}{\scalebox{0.76}{LogTrans}}}\\  
    \cmidrule(lr){2-3} \cmidrule(lr){4-5}\cmidrule(lr){6-7} \cmidrule(lr){8-9}\cmidrule(lr){10-11}\cmidrule(lr){12-13}\cmidrule(lr){14-15}\cmidrule(lr){16-17}\cmidrule(lr){18-19}\cmidrule(lr){20-21}\cmidrule(lr){22-23}
    \multicolumn{1}{c}{Metric} & \scalebox{0.76}{MSE} & \scalebox{0.76}{MAE} & \scalebox{0.76}{MSE} & \scalebox{0.76}{MAE} & \scalebox{0.76}{MSE} & \scalebox{0.76}{MAE} & \scalebox{0.76}{MSE} & \scalebox{0.76}{MAE} & \scalebox{0.76}{MSE} & \scalebox{0.76}{MAE} & \scalebox{0.76}{MSE} & \scalebox{0.76}{MAE} & \scalebox{0.76}{MSE} & \scalebox{0.76}{MAE} & \scalebox{0.76}{MSE} & \scalebox{0.76}{MAE} & \scalebox{0.76}{MSE} & \scalebox{0.76}{MAE} & \scalebox{0.76}{MSE} & \scalebox{0.76}{MAE} & \scalebox{0.76}{MSE} & \scalebox{0.76}{MAE}  \\
    \toprule
    \scalebox{0.76}{Electricity} &\boldres{\scalebox{0.76}{0.171}} &\boldres{\scalebox{0.76}{0.264}} &\secondres{\scalebox{0.76}{0.192}} &\secondres{\scalebox{0.76}{0.295}} & \scalebox{0.76}{0.208} & \scalebox{0.76}{0.323} & \scalebox{0.76}{0.229} &\scalebox{0.76}{0.329} &\scalebox{0.76}{0.212} &\scalebox{0.76}{0.300} &\scalebox{0.76}{0.214} &\scalebox{0.76}{0.327} &\scalebox{0.76}{0.193} &\scalebox{0.76}{0.296} &\scalebox{0.76}{0.227} &\scalebox{0.76}{0.338} &\scalebox{0.76}{0.379} &\scalebox{0.76}{0.445} &\scalebox{0.76}{0.311} &\scalebox{0.76}{0.397} &\scalebox{0.76}{0.272} &\scalebox{0.76}{0.370} \\
    \midrule
    \scalebox{0.76}{Traffic} &\boldres{\scalebox{0.76}{0.465}} &\boldres{\scalebox{0.76}{0.304}} &\scalebox{0.76}{0.620} &\secondres{\scalebox{0.76}{0.336}} & \scalebox{0.76}{0.621} & \scalebox{0.76}{0.396} & \scalebox{0.76}{0.622} &\scalebox{0.76}{0.392} &\scalebox{0.76}{0.625} &\scalebox{0.76}{0.383} &\secondres{\scalebox{0.76}{0.610}} &\scalebox{0.76}{0.376} &\scalebox{0.76}{0.624} &\scalebox{0.76}{0.340} &\scalebox{0.76}{0.628} &\scalebox{0.76}{0.379} &\scalebox{0.76}{0.878} &\scalebox{0.76}{0.469} &\scalebox{0.76}{0.764} &\scalebox{0.76}{0.416} &\scalebox{0.76}{0.705} &\scalebox{0.76}{0.395} \\
    \midrule
    \scalebox{0.76}{Weather} &\boldres{\scalebox{0.76}{0.249}} &\boldres{\scalebox{0.76}{0.275}} &\secondres{\scalebox{0.76}{0.259}} &\secondres{\scalebox{0.76}{0.287}} & \scalebox{0.76}{0.271} & \scalebox{0.76}{0.334} & \scalebox{0.76}{0.261} &\scalebox{0.76}{0.312} &\scalebox{0.76}{0.265} &\scalebox{0.76}{0.317} &\scalebox{0.76}{0.309} &\scalebox{0.76}{0.360} &\scalebox{0.76}{0.288} &\scalebox{0.76}{0.314} &\scalebox{0.76}{0.338} &\scalebox{0.76}{0.382} &\scalebox{0.76}{0.946} &\scalebox{0.76}{0.717} &\scalebox{0.76}{0.634} &\scalebox{0.76}{0.548} &\scalebox{0.76}{0.696} &\scalebox{0.76}{0.602} \\
    \midrule
    \scalebox{0.76}{ETTh1} &\secondres{\scalebox{0.76}{0.452}} & \boldres{\scalebox{0.76}{0.445}} &\scalebox{0.76}{0.458} & \secondres{\scalebox{0.76}{0.450}} & \scalebox{0.76}{0.542} & \scalebox{0.76}{0.510} & \scalebox{0.76}{0.491} &\scalebox{0.76}{0.479} &\scalebox{0.76}{0.456} &\scalebox{0.76}{0.452} &\boldres{\scalebox{0.76}{0.440}} &\scalebox{0.76}{0.460} &\scalebox{0.76}{0.570} &\scalebox{0.76}{0.537} &\scalebox{0.76}{0.496} &\scalebox{0.76}{0.487} &\scalebox{0.76}{0.827} &\scalebox{0.76}{0.703} &\scalebox{0.76}{1.040} &\scalebox{0.76}{0.795} &\scalebox{0.76}{1.072} &\scalebox{0.76}{0.837} \\
    \midrule
    \scalebox{0.76}{ETTh2} & \boldres{\scalebox{0.76}{0.386}} &\boldres{\scalebox{0.76}{0.411}} & \secondres{\scalebox{0.76}{0.414}} &\secondres{\scalebox{0.76}{0.427}} & \scalebox{0.76}{0.439} & \scalebox{0.76}{0.452} & \scalebox{0.76}{0.602} &\scalebox{0.76}{0.543} &\scalebox{0.76}{0.559} &\scalebox{0.76}{0.515} &\scalebox{0.76}{{0.437}} &\scalebox{0.76}{{0.449}} &\scalebox{0.76}{0.526} &\scalebox{0.76}{0.516} &\scalebox{0.76}{0.450} &\scalebox{0.76}{0.459} &\scalebox{0.76}{0.826} &\scalebox{0.76}{0.703} &\scalebox{0.76}{4.431} &\scalebox{0.76}{1.729} &\scalebox{0.76}{2.686} &\scalebox{0.76}{1.494} \\
    \midrule
    \scalebox{0.76}{ETTm1} &\boldres{\scalebox{0.76}{0.399}} &\boldres{\scalebox{0.76}{0.401}} &\secondres{\scalebox{0.76}{0.400}} &\secondres{\scalebox{0.76}{0.406}} & \scalebox{0.76}{0.429} & \scalebox{0.76}{0.425} & \scalebox{0.76}{0.435} &\scalebox{0.76}{0.437} &\scalebox{0.76}{0.403} &\scalebox{0.76}{0.407} &\scalebox{0.76}{0.448} &\scalebox{0.76}{0.452} &\scalebox{0.76}{0.481} &\scalebox{0.76}{0.456} &\scalebox{0.76}{0.588} &\scalebox{0.76}{0.517} &\scalebox{0.76}{0.691} &\scalebox{0.76}{0.607} &\scalebox{0.76}{0.961} &\scalebox{0.76}{0.734} &\scalebox{0.76}{0.929} &\scalebox{0.76}{0.725} \\
    \midrule
    \scalebox{0.76}{ETTm2} &\boldres{\scalebox{0.76}{0.291}} &\boldres{\scalebox{0.76}{0.330}} &\boldres{\scalebox{0.76}{0.291}} &\secondres{\scalebox{0.76}{0.333}} & \secondres{\scalebox{0.76}{0.293}} & \scalebox{0.76}{0.342} & \scalebox{0.76}{0.409} &\scalebox{0.76}{0.436} &\scalebox{0.76}{0.350} &\scalebox{0.76}{0.401} &\scalebox{0.76}{0.305} &\scalebox{0.76}{0.349} &\scalebox{0.76}{0.306} &\scalebox{0.76}{0.347} &\scalebox{0.76}{0.327} &\scalebox{0.76}{0.371} &\scalebox{0.76}{1.498} &\scalebox{0.76}{0.869} &\scalebox{0.76}{1.410} &\scalebox{0.76}{0.810} &\scalebox{0.76}{1.535} &\scalebox{0.76}{0.900} \\
    \midrule
    \scalebox{0.76}{Exchange} &\secondres{\scalebox{0.76}{0.355}} &\boldres{\scalebox{0.76}{0.403}} &\scalebox{0.76}{0.416} &\scalebox{0.76}{0.443} & \scalebox{0.76}{0.410} & \scalebox{0.76}{0.427} & \scalebox{0.76}{0.385} &\scalebox{0.76}{0.447} &\boldres{\scalebox{0.76}{0.354}} &\secondres{\scalebox{0.76}{0.414}} &\scalebox{0.76}{0.519} &\scalebox{0.76}{0.500} &\scalebox{0.76}{0.461} &\scalebox{0.76}{0.454} &\scalebox{0.76}{0.613} &\scalebox{0.76}{0.539} &\scalebox{0.76}{1.913} &\scalebox{0.76}{1.159} &\scalebox{0.76}{1.550} &\scalebox{0.76}{0.998} &\scalebox{0.76}{1.402} &\scalebox{0.76}{0.968} \\
    \midrule
    \scalebox{0.76}{ILI} &\boldres{\scalebox{0.76}{2.027}} &\boldres{\scalebox{0.76}{0.895}} &\scalebox{0.76}{2.139} &\scalebox{0.76}{0.931} & \scalebox{0.76}{2.497}& \scalebox{0.76}{1.004} & \scalebox{0.76}{7.382} &\scalebox{0.76}{2.003} &\scalebox{0.76}{2.616} &\scalebox{0.76}{1.090} &\scalebox{0.76}{2.847} &\scalebox{0.76}{1.144} &\secondres{\scalebox{0.76}{2.077}} &\secondres{\scalebox{0.76}{0.914}} &\scalebox{0.76}{3.006} &\scalebox{0.76}{1.161} &\scalebox{0.76}{7.635} &\scalebox{0.76}{2.050} &\scalebox{0.76}{5.137} &\scalebox{0.76}{1.544} &\scalebox{0.76}{4.839} &\scalebox{0.76}{1.485} \\
    \bottomrule
  \end{tabular}
    \end{small}
  \end{threeparttable}
  \vspace{-10pt}
\end{table}
\subsection{Ablation Studies}

In this section, we do some ablation experiments to show that our modifications of Transformer and the newly added modules are effective. To evaluate the function of the integrated linear module and ReVIN, we test the results of removing these modules from Client. To prove that addition embedding layer will cause a loss of information, we test the results of adding an embeding layer before the encoder blocks, and to demonstrate that a projection layer is better than a decoder, we test the result of replacing the projection layer with a decoder.

\begin{figure}[!htbp]
% \vspace{-15pt}
  \includegraphics[width=\columnwidth]{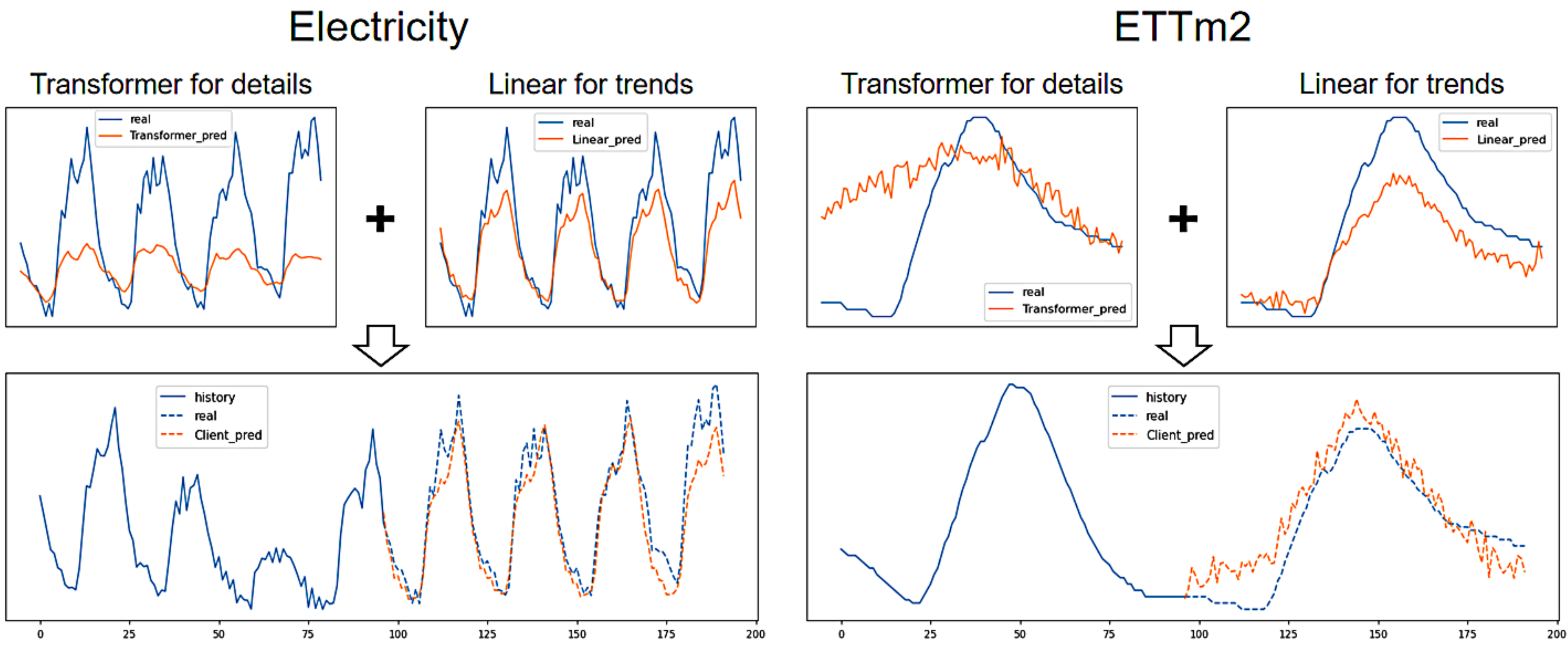}
  \centering
  \vspace{-5pt}
  \caption{Prediction showcases of Electricity and ETTm2. In each figure, the sub-figure in the top left represents the prediction of the Transformer module of Client, and the sub-figure in the top right represents the prediction of the linear module of Client. The sub-figure at the bottom shows the historical series and the final prediction of Client.}
  \label{fig:ecl}
  % \vspace{-15pt}
\end{figure}

\textbf{Integrated linear module} 
The integrated linear module is particularly effective at capturing trend information, while the Transformer module excels in capturing nonlinearities and cross-variable dependencies.
The LTSF results of removing the linear module from the Client are shown in the Table \ref{tab:ablation}. The performance of the Client without linear modules has declined on multiple datasets, especially on the ETTm1 and ILI datasets.
The prediction showcases of Client shown in Figure \ref{fig:ecl} illustrate that the linear module can already fit the long-term trends, while the Transformer module can supplement the short-term fluctuations. 

\textbf{ReVIN}
The ReVIN module, as presented in \cite{kim2022reversible}, plays a crucial role in addressing the problem of distribution shift. To investigate its impact on performance, we conduct ablation experiments by removing the ReVIN module from Client and report the results in Table \ref{tab:ablation}. Our findings suggest that except for the Electricity data which has a relatively stable distribution, the performance of the model has declined significantly on other datasets without ReVIN.

\textbf{Removing embedding layer} \label{para:embed}
Our model uses the time series data as the input of the Transformer without an embedding layer.
We speculate that introducing an additional embedding layer undermines inherent temporal information.
Table \ref{tab:ablation} presents the LTSF results of incorporating an embedding layer into the Client. It can be seen that the added embedding layer degrades the performance of the model on almost all datasets.

\textbf{Projection layer in place of decoder}
Our model replaces the decoder part with a simple projection layer. In our view, the Transformer module in Client functions more like a feature extractor rather than a series generator, and there is no temporal relationship among different variables. Consequently, we consider a decoder to be an unnecessary component in this context. Our ablation experiments, displayed in Table \ref{tab:ablation}, demonstrate that adding a decoder leads to a significant drop in the model's performance, as we anticipated.

\begin{table}[!htbp]
  \caption{Ablation experiments. The results of removing the linear module, removing the ReVIN module, adding an embedding layer, and replacing the projection layer with a decoder are listed in the table. Bold represents the best result.}
  \label{tab:ablation}
  \centering
  \begin{small}
  \renewcommand{\multirowsetup}{\centering}
  \setlength{\tabcolsep}{4.6pt}
  \begin{tabular}{c|c|cc|cc|cc|cc|cc}
    \toprule
    \multicolumn{2}{c}{Models} &
    \multicolumn{2}{c}{Client} &
    \multicolumn{2}{c}{Client - Linear} &
    \multicolumn{2}{c}{Client - ReVIN} &
    \multicolumn{2}{c}{Client + Embed} &
    \multicolumn{2}{c}{Client + Decoder} 
    \\
    \cmidrule(lr){3-4} 
    \cmidrule(lr){5-6}
    \cmidrule(lr){7-8} 
    \cmidrule(lr){9-10}
    \cmidrule(lr){11-12}
    \multicolumn{2}{c}{Metric} & MSE & MAE & MSE & MAE & MSE & MAE & MSE & MAE & MSE & MAE  \\
    \toprule
    \multirow{4}{*}{\rotatebox{90}{ETTm1}} 
    & 96  & \textbf{0.336} & \textbf{0.369} & 0.348 & 0.378  & 0.394 & 0.430 & 0.339 & 0.375 & 0.689 & 0.549\\
    & 192 & \textbf{0.376} & \textbf{0.385} & 0.386 & 0.395  & 0.441 & 0.458 & 0.384 & 0.396 & 0.705 & 0.557\\
    & 336 & \textbf{0.408} & \textbf{0.407} & 0.419 & 0.416  & 0.490 & 0.486 & 0.425 & 0.426 & 0.706 & 0.555\\
    & 720 & \textbf{0.477} & \textbf{0.442} & 0.484 & 0.451  & 0.540 & 0.516 & 0.495 & 0.464 & 0.737 & 0.575\\
    \midrule
    \multirow{4}{*}{\rotatebox{90}{ILI}} 
    & 24  & \textbf{2.033} & \textbf{0.870} & 2.934 & 1.013  & 4.110 & 1.400 & 2.650 & 1.030 & 4.518 & 1.412\\
    & 36  & \textbf{1.909} & \textbf{0.868} & 2.355 & 0.974  & 4.340 & 1.444 & 2.490 & 0.982 & 4.328 & 1.394\\
    & 48  & \textbf{2.126} & \textbf{0.929} & 2.341 & 0.976  & 4.330 & 1.427 & 2.504 & 0.994 & 4.615 & 1.436\\
    & 60  & \textbf{2.039} & \textbf{0.914} & 2.385 & 0.968  & 4.528 & 1.476 & 2.505 & 0.998 & 4.464 & 1.432\\
    \midrule
    \multirow{4}{*}{\rotatebox{90}{Electricity}} 
     & 96 & \textbf{0.141} & \textbf{0.236} & 0.143 & 0.239 & 0.147 & 0.244 & 0.165 & 0.264 & 0.206 & 0.297\\
    & 192 & \textbf{0.161} & 0.254 & \textbf{0.161} & \textbf{0.253} & 0.163 & 0.262 & 0.179 & 0.282 & 0.209 & 0.298\\
    & 336 & \textbf{0.173} & \textbf{0.267} & 0.179 & 0.273 & 0.176 & 0.279 & 0.194 & 0.293 & 0.215 & 0.307\\
    & 720 & 0.209 & \textbf{0.299} & 0.210 & 0.301 & \textbf{0.206} & 0.309 & 0.210 & 0.304 & 0.284 & 0.364\\
    \midrule
    \multirow{4}{*}{\rotatebox{90}{Traffic}} 
    & 96  & \textbf{0.438} & \textbf{0.292} & 0.442 & 0.301  & 0.591 & 0.384 & 0.448 & 0.300 & 0.643 & 0.391\\
    & 192 & \textbf{0.451} & \textbf{0.298} & 0.453 & 0.303  & 0.586 & 0.391 & 0.481 & 0.325 & 0.598 & 0.363\\
    & 336 & 0.472 & \textbf{0.305} & \textbf{0.471} & 0.306  & 0.593 & 0.394 & 0.497 & 0.332 & 0.608 & 0.369\\
    & 720 & \textbf{0.499} & \textbf{0.321} & 0.502 & 0.324  & 0.645 & 0.408 & 0.542 & 0.365 & 0.645 & 0.396\\
    \midrule
    \multirow{4}{*}{\rotatebox{90}{ETTh1}} 
    & 96  & \textbf{0.392} & 0.409 & 0.396 & 0.408  & 0.448 & 0.466 & 0.397 & \textbf{0.405} & 0.763 & 0.605\\
    & 192 & 0.445 & 0.436 & \textbf{0.441} & \textbf{0.435}  & 0.522 & 0.512 & 0.467 & 0.447 & 0.845 & 0.653\\
    & 336 & \textbf{0.482} & \textbf{0.455} & 0.491 & 0.462  & 0.540 & 0.514 & 0.504 & 0.469 & 0.951 & 0.709\\
    & 720 & \textbf{0.489} & \textbf{0.479} & 0.492 & 0.482  & 0.653 & 0.605 & 0.608 & 0.545 & 0.911 & 0.716\\
    \bottomrule
  \end{tabular}
  \end{small}
\end{table}

\subsection{Additional analysis of Client}

\textbf{Why Client is more effective than linear models?} 
In our view, Client is more effective than linear models for two main reasons. Firstly, Client can capture the strong correlations between series of different variables thanks to the cross-variable Transformer module. Secondly, Client is capable of fitting complex nonlinearities within a series, which allows it to effectively capture nonlinear features that are often present in real-world time series data.
However, the effectiveness of the Client model is more obvious on the datasets with with abundant nonlinearities and cross-variable dependencies, such as the Electricity and Traffic datasets. A further variable correlation analysis is shown in Appendix \ref{app:corr}.

\textbf{Client Forecasting results with different look-back window sizes}
According to DLinear \cite{Zeng2022AreTE}, the forecasting abilities of the previous Transformer models don't improve significantly with an increase in the size of the input's look-back window. This observation is not surprising because, based on our experiments, previous cross-time based Transformer models fail to fully learn even when the series has a length of $96$, indicating that increasing the historical input of the model may not be effective. As demonstrated in Table \ref{tab:ts_extend}, Client's performance improves as the size of the look-back window increases. We also test the previous SOTA model TimesNet, but find that its ability does not improve significantly as the look-back window grows.

\begin{table}[htbp]
  \caption{Forecasting results as the look-back window size grows. For Client and TimesNet, three different look-back window sizes are compared, which are 96 (the basic setting for all experiments), 144, and 192. The number added after the model represents the look-back window size, and bold represents the best result.}
  \label{tab:ts_extend}
  \centering
  \begin{small}
  \renewcommand{\multirowsetup}{\centering}
  \setlength{\tabcolsep}{4.2pt}
  \begin{tabular}{c|c|cc|cc|cc|cc|cc|cc}
    \toprule
    \multicolumn{2}{c}{Models} &
    \multicolumn{2}{c}{Client + 96} &
    \multicolumn{2}{c}{Client + 144} &
    \multicolumn{2}{c}{Client + 192} &
    \multicolumn{2}{c}{TimesNet + 96} &
    \multicolumn{2}{c}{TimesNet + 144} &
    \multicolumn{2}{c}{TimesNet + 192}
    \\
    \cmidrule(lr){3-4} 
    \cmidrule(lr){5-6}
    \cmidrule(lr){7-8} 
    \cmidrule(lr){9-10}
    \cmidrule(lr){11-12}
    \cmidrule(lr){13-14}
    \multicolumn{2}{c}{Metric} & MSE & MAE & MSE & MAE & MSE & MAE & MSE & MAE & MSE & MAE & MSE & MAE \\
    \toprule
    \multirow{4}{*}{\rotatebox{90}{Electricity}} 
    & 96  & 0.141 & 0.236 & 0.134 & 0.229  & \textbf{0.132} & \textbf{0.227} & 0.168 & 0.272 & \textbf{0.166} & \textbf{0.271} & 0.171 & 0.275\\
    & 192 & 0.161 & 0.254 & 0.155 & 0.247  & \textbf{0.151} & \textbf{0.244} & 0.184 & 0.289 & \textbf{0.182} & \textbf{0.284} & 0.185 & 0.288\\
    & 336 & 0.173 & 0.267 & 0.171 & 0.264  & \textbf{0.167} & \textbf{0.261} & 0.198 & 0.300 & \textbf{0.192} & 0.296 & 0.193 & \textbf{0.295}  \\
    & 720 & 0.209 & 0.299 & 0.208 & 0.298  & \textbf{0.198} & \textbf{0.289} & \textbf{0.220} & \textbf{0.320} & 0.237 & 0.346 & 0.285 & 0.362  \\
    \bottomrule
  \end{tabular}
  \end{small}
\end{table}

\begin{table}[htbp]
  \caption{The experiments of replacing the attention module. "Attention" means the typical attention module, which is used in Client. "ProbAttention", "Linear" and "MLP" means replacing the the typical attention module with these modules, and "No Attention" means removing the attention module from Client. Bold represents the best result.}
  \label{tab:replace-attention}
  \centering
  \begin{small}
  \renewcommand{\multirowsetup}{\centering}
  \setlength{\tabcolsep}{4.6pt}
  \begin{tabular}{c|c|cc|cc|cc|cc|cc}
    \toprule
    \multicolumn{2}{c}{Models} &
    \multicolumn{2}{c}{Attention} &
    \multicolumn{2}{c}{ProbAttention} &
    \multicolumn{2}{c}{Linear} &
    \multicolumn{2}{c}{MLP} &
    \multicolumn{2}{c}{No Attention}
    \\
    \cmidrule(lr){3-4} 
    \cmidrule(lr){5-6}
    \cmidrule(lr){7-8} 
    \cmidrule(lr){9-10}
    \cmidrule(lr){11-12}
    \multicolumn{2}{c}{Metric} & MSE & MAE & MSE & MAE & MSE & MAE & MSE & MAE & MSE & MAE \\
    \toprule
    \multirow{4}{*}{\rotatebox{90}{Electricity}} 
    & 96  & \textbf{0.141} & \textbf{0.236} & 0.143 & 0.237  & 0.166 & 0.266 & 0.158 & 0.257 & 0.160 & 0.250\\
    & 192 & 0.161 & 0.254 & \textbf{0.159} & \textbf{0.252}  & 0.177 & 0.275 & 0.173 & 0.269 & 0.171 & 0.260\\
    & 336 & \textbf{0.173} & \textbf{0.267} & 0.175 & 0.268  & 0.196 & 0.290 & 0.189 & 0.285 & 0.188 & 0.277\\
    & 720 & \textbf{0.209} & 0.299 & \textbf{0.209} & \textbf{0.295}  & 0.216 & 0.309 & 0.217 & 0.312 & 0.228 & 0.311\\
    \bottomrule
  \end{tabular}
  \end{small}
\end{table}

% \begin{wraptable}{r}{8.5cm}

\textbf{Replacing the attention module}
The attention module has been demonstrated effective in capturing the dependencies between different variables. We also test capturing the dependencies between variables using other modules, including ProbAttention used in Informer \cite{haoyietal-informer-2021}, linear layer and MLP layer. 
We also test the performance of removing the attention module from Client, which is equivalent to only adding non-linear expression ability to the linear model.
The results of these experiments are presented in the Table \ref{tab:replace-attention}. 
It can be observed that the models with the typical attention and ProbAttention have better forecasting results than the ones with the linear layer or MLP layer, or without the attention module.

\textbf{Improved efficiency of model training}
Similar to other Transformer-based models, the most computational part of Client is the calculation of attention. As Client's attention module is used to learn dependencies between variables in multivariate time series, its complexity is $O(N^2)$, where $N$ is the number of variables and can be reduced to $O(N\log N)$ by using ProbSparse~\cite{haoyietal-informer-2021}. 

Nevertheless, Client removes the Decoder component, the embedding layer and the positional encoding layer, effectively halving the training time and memory consumption.
Table \ref{tab:efficiency} shows the efficiency comparison of Client, TimesNet, and vanilla Transformer.
Client also outperforms vanilla Transformer and TimesNet in terms of parameter quantity and GPU memory usage during training.

\begin{table}
  \caption{Model efficiency comparison. Model efficiency is compared in three aspects: model parameter quantity (M), GPU memory usage during training (MiB), and time per iteration (seconds). Client is compared with TimesNet and vanilla Transformer. The look-back window and prediction window length are 96 (ILI 36, 24).}
  \label{tab:efficiency}
  \vskip 0.05in
  \centering
  \begin{threeparttable}
  \begin{small}
  \renewcommand{\multirowsetup}{\centering}
  \setlength{\tabcolsep}{0.3pt}
  \begin{tabular}{c|ccc|ccc|ccc}
    \toprule
    \multicolumn{1}{c}{\multirow{2}{*}{Models}} & 
    \multicolumn{3}{c}{\rotatebox{0}{\scalebox{0.76}{\textbf{Client}}}} &
    \multicolumn{3}{c}{\rotatebox{0}{\scalebox{0.76}{TimesNet}}} &
    \multicolumn{3}{c}{\rotatebox{0}{\scalebox{0.76}{Transformer}}} \\
    \cmidrule(lr){2-4} \cmidrule(lr){5-7}\cmidrule(lr){8-10}
    \multicolumn{1}{c}{Metric} & \scalebox{0.76}{Parameter} & \scalebox{0.76}{Memory} & \scalebox{0.76}{Time} & \scalebox{0.76}{Parameter} & \scalebox{0.76}{Memory} & \scalebox{0.76}{Time} & \scalebox{0.76}{Parameter} & \scalebox{0.76}{Memory} & \scalebox{0.76}{Time}  \\
    \toprule
    \scalebox{0.76}{Electricity} &\textbf{\scalebox{0.76}{0.886}} &\scalebox{0.76}{3008} &\textbf{\scalebox{0.76}{0.017}} &\scalebox{0.76}{150.305} & \scalebox{0.76}{12012} & \scalebox{0.76}{0.487} & \scalebox{0.76}{11.666} &\textbf{\scalebox{0.76}{2898}} &\scalebox{0.76}{0.019}   \\
    \midrule
    \scalebox{0.76}{Traffic} &\textbf{\scalebox{0.76}{0.294}} &\scalebox{0.76}{7606} &\scalebox{0.76}{0.056} &\scalebox{0.76}{301.693} & \scalebox{0.76}{10182} & \scalebox{0.76}{0.940} & \scalebox{0.76}{8.882} &\textbf{\scalebox{0.76}{2570}} &\textbf{\scalebox{0.76}{0.026}} \\
    \midrule
    \scalebox{0.76}{Weather} &\textbf{\scalebox{0.76}{0.107}} &\textbf{\scalebox{0.76}{1832}} &\textbf{\scalebox{0.76}{0.006}} &\scalebox{0.76}{1.194} & \scalebox{0.76}{2386} & \scalebox{0.76}{0.032} & \scalebox{0.76}{10.590} &\scalebox{0.76}{2859} &\scalebox{0.76}{0.015} \\
    \midrule
    \scalebox{0.76}{ETTh1 \& ETTh2} &\textbf{\scalebox{0.76}{0.107}} & \textbf{\scalebox{0.76}{1824}} &\textbf{\scalebox{0.76}{0.006}} & \scalebox{0.76}{0.605} & \scalebox{0.76}{2300} & \scalebox{0.76}{0.030} & \scalebox{0.76}{10.540} &\scalebox{0.76}{2854} &\scalebox{0.76}{0.015} \\
    \midrule
    \scalebox{0.76}{ETTm1 \& ETTm2} &\textbf{{\scalebox{0.76}{0.119}}} & \textbf{{\scalebox{0.76}{1804}}} &\textbf{\scalebox{0.76}{0.006}} & \scalebox{0.76}{4.708} & \scalebox{0.76}{2216} & \scalebox{0.76}{0.036} & \scalebox{0.76}{10.540} &\scalebox{0.76}{2854} &\scalebox{0.76}{0.015} \\
    \midrule
    \scalebox{0.76}{Exchange} &\textbf{{\scalebox{0.76}{0.885}}} & \textbf{\scalebox{0.76}{1956}} &\textbf{\scalebox{0.76}{0.040}} & \scalebox{0.76}{4.708} & \scalebox{0.76}{2356} & \scalebox{0.76}{0.245} & \scalebox{0.76}{10.544} &\scalebox{0.76}{2854} &\scalebox{0.76}{0.077} \\
    \midrule
    \scalebox{0.76}{ILI} &\textbf{\scalebox{0.76}{0.311}} & \textbf{\scalebox{0.76}{1936}} &\textbf{\scalebox{0.76}{0.005}} & \scalebox{0.76}{674.805} & \scalebox{0.76}{18024} & \scalebox{0.76}{0.773} & \scalebox{0.76}{10.540} &\scalebox{0.76}{2520} &\scalebox{0.76}{0.009} \\
    \bottomrule
  \end{tabular}
    \end{small}
  \end{threeparttable}
\end{table}

\section{Conclusion}
This paper proposes Client, a Cross-variable Linear Integrated Enhanced Transformer for Multivariate Long-Term Time Series.
By using a cross-variable Transformer in place of the cross-time Transformer, this paper dispel the doubt on the effectiveness of Transformer model for time series forecasting, and it demonstrates that cross-variable attention in Transformer is more important than cross-time attention for LTSF. With the linear model integration and other model modifications, Client achieves the state-of-the-art performance on multiple real-world datasets with the least training time and GPU memory consumption compared with the previous Transformer models or TimesNet. Furthermore, Client exhibits better forecasting results as the size of the look-back window increases. The mask-series experiments conducted in this study also highlight the limitations of many previous cross-time based Transformer models in effectively capturing information from time series data. 
Therefore, the proposed Client model provides a valuable solution that addresses this limitation and paves the way for significant improvements in time series forecasting performance.

\small
\bibliographystyle{plain}
\bibliography{neurips_2023}

\begin{thebibliography}{10}

\bibitem{box2015time}
George~EP Box, Gwilym~M Jenkins, Gregory~C Reinsel, and Greta~M Ljung.
\newblock {\em Time series analysis: forecasting and control}.
\newblock John Wiley \& Sons, 2015.

\bibitem{brown2020language}
Tom Brown, Benjamin Mann, Nick Ryder, Melanie Subbiah, Jared~D Kaplan, Prafulla
  Dhariwal, Arvind Neelakantan, Pranav Shyam, Girish Sastry, Amanda Askell,
  et~al.
\newblock Language models are few-shot learners.
\newblock {\em Advances in neural information processing systems},
  33:1877--1901, 2020.

\bibitem{ilidata}
{CDC}.
\newblock {Illness}.
\newblock \url{https://gis.cdc.gov/grasp/fluview/fluportaldashboard.html}.

\bibitem{cui2020calibrated}
Peng Cui, Wenbo Hu, and Jun Zhu.
\newblock Calibrated reliable regression using maximum mean discrepancy.
\newblock {\em Advances in Neural Information Processing Systems},
  33:17164--17175, 2020.

\bibitem{devlin2018bert}
Jacob Devlin, Ming-Wei Chang, Kenton Lee, and Kristina Toutanova.
\newblock Bert: Pre-training of deep bidirectional transformers for language
  understanding.
\newblock {\em arXiv preprint arXiv:1810.04805}, 2018.

\bibitem{fang2022attention}
Weiwei Fang, Wenhao Zhuo, Jingwen Yan, Youyi Song, Dazhi Jiang, and Teng Zhou.
\newblock Attention meets long short-term memory: A deep learning network for
  traffic flow forecasting.
\newblock {\em Physica A: Statistical Mechanics and its Applications},
  587:126485, 2022.

\bibitem{gao2023adaptive}
Jiaxin Gao, Yuntian Chen, Wenbo Hu, and Dongxiao Zhang.
\newblock An adaptive deep-learning load forecasting framework by integrating
  transformer and domain knowledge.
\newblock {\em Advances in Applied Energy}, page 100142, 2023.

\bibitem{gulati2020conformer}
Anmol Gulati, James Qin, Chung-Cheng Chiu, Niki Parmar, Yu~Zhang, Jiahui Yu,
  Wei Han, Shibo Wang, Zhengdong Zhang, Yonghui Wu, et~al.
\newblock Conformer: Convolution-augmented transformer for speech recognition.
\newblock {\em arXiv preprint arXiv:2005.08100}, 2020.

\bibitem{Hochreiter1997LongSM}
S.~Hochreiter and J.~Schmidhuber.
\newblock Long short-term memory.
\newblock {\em Neural Comput.}, 1997.

\bibitem{kim2022reversible}
Taesung Kim, Jinhee Kim, Yunwon Tae, Cheonbok Park, Jang-Ho Choi, and Jaegul
  Choo.
\newblock Reversible instance normalization for accurate time-series
  forecasting against distribution shift.
\newblock In {\em ICLR}, 2022.

\bibitem{DBLP:journals/corr/KingmaB14}
Diederik~P. Kingma and Jimmy Ba.
\newblock Adam: {A} method for stochastic optimization.
\newblock In {\em ICLR}, 2015.

\bibitem{2018Modeling}
Guokun Lai, Wei-Cheng Chang, Yiming Yang, and Hanxiao Liu.
\newblock Modeling long-and short-term temporal patterns with deep neural
  networks.
\newblock In {\em SIGIR}, 2018.

\bibitem{1998Gradient}
Y.~Lecun and L.~Bottou.
\newblock Gradient-based learning applied to document recognition.
\newblock {\em Proceedings of the IEEE}, 86(11):2278--2324, 1998.

\bibitem{2019Enhancing}
Shiyang Li, Xiaoyong Jin, Yao Xuan, Xiyou Zhou, Wenhu Chen, Yu-Xiang Wang, and
  Xifeng Yan.
\newblock Enhancing the locality and breaking the memory bottleneck of
  transformer on time series forecasting.
\newblock In {\em NeurIPS}, 2019.

\bibitem{liu2022scinet}
Minhao Liu, Ailing Zeng, Muxi Chen, Zhijian Xu, Qiuxia Lai, Lingna Ma, and
  Qiang Xu.
\newblock Scinet: Time series modeling and forecasting with sample convolution
  and interaction.
\newblock {\em Advances in Neural Information Processing Systems},
  35:5816--5828, 2022.

\bibitem{liu2021pyraformer}
Shizhan Liu, Hang Yu, Cong Liao, Jianguo Li, Weiyao Lin, Alex~X Liu, and
  Schahram Dustdar.
\newblock Pyraformer: Low-complexity pyramidal attention for long-range time
  series modeling and forecasting.
\newblock In {\em ICLR}, 2021.

\bibitem{Liu2022NonstationaryTR}
Yong Liu, Haixu Wu, Jianmin Wang, and Mingsheng Long.
\newblock Non-stationary transformers: Rethinking the stationarity in time
  series forecasting.
\newblock In {\em NeurIPS}, 2022.

\bibitem{liu2021swin}
Ze~Liu, Yutong Lin, Yue Cao, Han Hu, Yixuan Wei, Zheng Zhang, Stephen Lin, and
  Baining Guo.
\newblock Swin transformer: Hierarchical vision transformer using shifted
  windows.
\newblock In {\em Proceedings of the IEEE/CVF international conference on
  computer vision}, pages 10012--10022, 2021.

\bibitem{lopez2023can}
Alejandro Lopez-Lira and Yuehua Tang.
\newblock Can chatgpt forecast stock price movements? return predictability and
  large language models.
\newblock {\em arXiv preprint arXiv:2304.07619}, 2023.

\bibitem{meenal2022weather}
R~Meenal, D~Binu, KC~Ramya, Prawin~Angel Michael, K~Vinoth~Kumar,
  E~Rajasekaran, and B~Sangeetha.
\newblock Weather forecasting for renewable energy system: a review.
\newblock {\em Archives of Computational Methods in Engineering},
  29(5):2875--2891, 2022.

\bibitem{1991Multilayer}
F.~Murtagh.
\newblock Multilayer perceptrons for classification and regression.
\newblock {\em Neurocomputing}, 2(5–6):183--197, 1991.

\bibitem{trafficdata}
{PeMS}.
\newblock {Traffic}.
\newblock \url{http://pems.dot.ca.gov/}.

\bibitem{Flunkert2017DeepARPF}
David Salinas, Valentin Flunkert, Jan Gasthaus, and Tim Januschowski.
\newblock Deep{AR}: Probabilistic forecasting with autoregressive recurrent
  networks.
\newblock {\em Int. J. Forecast.}, 2020.

\bibitem{2009The}
F.~Scarselli, M.~Gori, A.~C. Tsoi, M.~Hagenbuchner, and G.~Monfardini.
\newblock The graph neural network model.
\newblock {\em IEEE Transactions on Neural Networks}, 20(1):61, 2009.

\bibitem{Sen2019ThinkGA}
Rajat Sen, Hsiang-Fu Yu, and Inderjit~S. Dhillon.
\newblock Think globally, act locally: A deep neural network approach to
  high-dimensional time series forecasting.
\newblock In {\em NeurIPS}, 2019.

\bibitem{ecldata}
{UCI}.
\newblock {Electricity}.
\newblock
  \url{https://archive.ics.uci.edu/ml/datasets/ElectricityLoadDiagrams20112014}.

\bibitem{Oord2016WaveNetAG}
A{\"a}ron van~den Oord, S.~Dieleman, H.~Zen, K.~Simonyan, Oriol Vinyals,
  A.~Graves, Nal Kalchbrenner, A.~Senior, and K.~Kavukcuoglu.
\newblock Wavenet: A generative model for raw audio.
\newblock In {\em SSW}, 2016.

\bibitem{2017attention}
Ashish Vaswani, Noam Shazeer, Niki Parmar, Jakob Uszkoreit, Llion Jones,
  Aidan~N Gomez, Lukasz Kaiser, and Illia Polosukhin.
\newblock Attention is all you need.
\newblock In {\em NeurIPS}, 2017.

\bibitem{wen2022transformers}
Qingsong Wen, Tian Zhou, Chaoli Zhang, Weiqi Chen, Ziqing Ma, Junchi Yan, and
  Liang Sun.
\newblock Transformers in time series: A survey.
\newblock {\em arXiv preprint arXiv:2202.07125}, 2022.

\bibitem{weatherdata}
{Wetterstation}.
\newblock {Weather}.
\newblock \url{https://www.bgc-jena.mpg.de/wetter/}.

\bibitem{woo2022etsformer}
Gerald Woo, Chenghao Liu, Doyen Sahoo, Akshat Kumar, and Steven C.~H. Hoi.
\newblock Etsformer: Exponential smoothing transformers for time-series
  forecasting.
\newblock {\em arXiv preprint arXiv:2202.01381}, 2022.

\bibitem{wu2023timesnet}
Haixu Wu, Tengge Hu, Yong Liu, Hang Zhou, Jianmin Wang, and Mingsheng Long.
\newblock Timesnet: Temporal 2d-variation modeling for general time series
  analysis.
\newblock In {\em International Conference on Learning Representations}, 2023.

\bibitem{wu2021autoformer}
Haixu Wu, Jiehui Xu, Jianmin Wang, and Mingsheng Long.
\newblock Autoformer: Decomposition transformers with {Auto-Correlation} for
  long-term series forecasting.
\newblock In {\em NeurIPS}, 2021.

\bibitem{wu2020connecting}
Zonghan Wu, Shirui Pan, Guodong Long, Jing Jiang, Xiaojun Chang, and Chengqi
  Zhang.
\newblock Connecting the dots: Multivariate time series forecasting with graph
  neural networks.
\newblock In {\em Proceedings of the 26th ACM SIGKDD international conference
  on knowledge discovery \& data mining}, pages 753--763, 2020.

\bibitem{Zeng2022AreTE}
Ailing Zeng, Muxi Chen, Lei Zhang, and Qiang Xu.
\newblock Are transformers effective for time series forecasting?
\newblock 2023.

\bibitem{Zhang2022LessIM}
T.~Zhang, Yizhuo Zhang, Wei Cao, J.~Bian, Xiaohan Yi, Shun Zheng, and Jian Li.
\newblock Less is more: Fast multivariate time series forecasting with light
  sampling-oriented mlp structures.
\newblock {\em arXiv preprint arXiv:2207.01186}, 2022.

\bibitem{zhang2023crossformer}
Yunhao Zhang and Junchi Yan.
\newblock Crossformer: Transformer utilizing cross-dimension dependency for
  multivariate time series forecasting.
\newblock In {\em The Eleventh International Conference on Learning
  Representations}, 2023.

\bibitem{zhang2020dynamic}
Zhiheng Zhang, Wenbo Hu, Tian Tian, and Jun Zhu.
\newblock Dynamic window-level granger causality of multi-channel time series.
\newblock {\em arXiv preprint arXiv:2006.07788}, 2020.

\bibitem{haoyietal-informer-2021}
Haoyi Zhou, Shanghang Zhang, Jieqi Peng, Shuai Zhang, Jianxin Li, Hui Xiong,
  and Wancai Zhang.
\newblock Informer: Beyond efficient transformer for long sequence time-series
  forecasting.
\newblock In {\em AAAI}, 2021.

\bibitem{zhou2022fedformer}
Tian Zhou, Ziqing Ma, Qingsong Wen, Xue Wang, Liang Sun, and Rong Jin.
\newblock {FEDformer}: Frequency enhanced decomposed transformer for long-term
  series forecasting.
\newblock In {\em ICML}, 2022.

\end{thebibliography}

\clearpage
\appendix
\begin{center}
\huge  Supplemental Materials for  
\par
\Large Client: Cross-variable Linear Integrated Enhanced Transformer for Multivariate Long-Term Time Series Forecasting
\end{center}

\section{Detailed analysis of Crossformer} \label{app:crossformer}
Crossformer leverages the Dimension-Segment-Wise (DSW) embedding to convert a single time series into a 2D vector and utilizes a Two-Stage Attention (TSA) layer to capture cross-time dependencies and dependencies of series segments across different dimensions. However, the slicing and embedding operations performed by Crossformer on the input series may lead to a loss of information (the experiments in \ref{para:embed} show that an extra embedding layer to encode time series results in a decrease in the model’s performance). Furthermore, the cross-time attention in Crossformer may have difficulty precisely capturing trend information in time series data, as demonstrated in the mask-series experiment, resulting in inferior performance compared to linear models across most datasets. In addition, the model's complex structure contributes to long training times and high memory requirements, which have been observed to cause out-of-memory (OOM) errors when predicting long series with many variables on budget-friendly GPUs.

\section{Data Description} \label{app:describe}
The experimental data includes 9 popular datasets with different characteristics:

Electricity: The dataset contains the hourly electricity consumption (Kwh) of 321 clients from 2012 to 2014. There are 321 variables with 26,304 timesteps per variable. 

Traffic: This dataset comprises hourly data for San Francisco freeways from 2015 to 2016. There are 862 variables with 17,544 timesteps per variable.

Weather: The dataset includes 21 types of weather data recorded every 10 minutes for the year 2020 in Germany. There are 21 variables with 52,696 timesteps per variable.

ETT: The ETT dataset comprises two hourly-level datasets (ETTh) and two 15-minute-level datasets (ETTm). Each dataset contains seven oil and load features of electricity transformers from July 2016 to July 2018. There are 7 variables for each dataset, and each variable has 17,420 timesteps for ETTh and 69,680 timesteps for ETTm.

Exchange-Rate: The dataset contains daily exchange rates of eight countries from 1990 to 2016. There are 8 variables with 7,588 timesteps per variable.

ILI: The dataset includes weekly data from the Centers for Disease Control and Prevention of the United States from 2002 to 2021. There are 7 variables with 966 timesteps per variable.

\section{Complete results} \label{app:complete}
The complete results for LTSF are shown in Table \ref{tab:com_forecasting_results}. The average of the results of 4 different prediction lengths and the number of optimal values obtained by different models are also listed in the table. Client achieves the best results in over 70\% of experimental settings.
\begin{table}[!htbp]
  \caption{The complete results for LTSF. The results of 4 different prediction lengths of different models are listed in the table. The look-back window sizes are set to 36 for ILI and 96 for the other datasets. We also calculate the average of the results for the 4 prediction lengths and the number of optimal values obtained by different models.}\label{tab:com_forecasting_results}
  \vskip 0.05in
  \centering
  \resizebox{1.1\columnwidth}{!}{
  \begin{threeparttable}
  \begin{small}
  \renewcommand{\multirowsetup}{\centering}
  \setlength{\tabcolsep}{1pt}
  \begin{tabular}{c|c|cc|cc|cc|cc|cc|cc|cc|cc|cc|cc|cc}
    \toprule
    \multicolumn{2}{c}{\multirow{2}{*}{Models}} & 
    \multicolumn{2}{c}{\rotatebox{0}{\scalebox{0.8}{\textbf{Client}}}} &
    \multicolumn{2}{c}{\rotatebox{0}{\scalebox{0.8}{TimesNet}}} &
    \multicolumn{2}{c}{\rotatebox{0}{\scalebox{0.8}{\update{ETSformer}}}} &
    \multicolumn{2}{c}{\rotatebox{0}{\scalebox{0.8}{LightTS$^\ast$}}} &
    \multicolumn{2}{c}{\rotatebox{0}{\scalebox{0.8}{DLinear$^\ast$}}} &
    \multicolumn{2}{c}{\rotatebox{0}{\scalebox{0.8}{FEDformer}}} & \multicolumn{2}{c}{\rotatebox{0}{\scalebox{0.8}{Stationary}}} & \multicolumn{2}{c}{\rotatebox{0}{\scalebox{0.8}{Autoformer}}} & \multicolumn{2}{c}{\rotatebox{0}{\scalebox{0.8}{Pyraformer}}} &  \multicolumn{2}{c}{\rotatebox{0}{\scalebox{0.8}{Informer}}} & \multicolumn{2}{c}{\rotatebox{0}{\scalebox{0.8}{LogTrans}}} \\
    \cmidrule(lr){3-4} \cmidrule(lr){5-6}\cmidrule(lr){7-8} \cmidrule(lr){9-10}\cmidrule(lr){11-12}\cmidrule(lr){13-14}\cmidrule(lr){15-16}\cmidrule(lr){17-18}\cmidrule(lr){19-20}\cmidrule(lr){21-22}\cmidrule(lr){23-24}
    \multicolumn{2}{c}{Metric} & \scalebox{0.78}{MSE} & \scalebox{0.78}{MAE} & \scalebox{0.78}{MSE} & \scalebox{0.78}{MAE} & \scalebox{0.78}{MSE} & \scalebox{0.78}{MAE} & \scalebox{0.78}{MSE} & \scalebox{0.78}{MAE} & \scalebox{0.78}{MSE} & \scalebox{0.78}{MAE} & \scalebox{0.78}{MSE} & \scalebox{0.78}{MAE} & \scalebox{0.78}{MSE} & \scalebox{0.78}{MAE} & \scalebox{0.78}{MSE} & \scalebox{0.78}{MAE} & \scalebox{0.78}{MSE} & \scalebox{0.78}{MAE} & \scalebox{0.78}{MSE} & \scalebox{0.78}{MAE} & \scalebox{0.78}{MSE} & \scalebox{0.78}{MAE}\\
    \toprule
    \multirow{5}{*}{\rotatebox{90}{\scalebox{0.95}{Electricity}}} 
    &  \scalebox{0.78}{96} &\boldres{\scalebox{0.78}{0.141}} &\boldres{\scalebox{0.78}{0.236}} &\scalebox{0.78}{0.168} &\scalebox{0.78}{0.272} & \scalebox{0.78}{0.187}& \scalebox{0.78}{0.304} &\scalebox{0.78}{0.207} &\scalebox{0.78}{0.307} &\scalebox{0.78}{0.197} &\scalebox{0.78}{0.282} &\scalebox{0.78}{0.193} &\scalebox{0.78}{0.308} &\secondres{\scalebox{0.78}{0.169}} &\secondres{\scalebox{0.78}{0.273}} &\scalebox{0.78}{0.201} &\scalebox{0.78}{0.317} &\scalebox{0.78}{0.386} &\scalebox{0.78}{0.449} &\scalebox{0.78}{0.274} &\scalebox{0.78}{0.368} &\scalebox{0.78}{0.258} &\scalebox{0.78}{0.357}  \\
    & \scalebox{0.78}{192} &\boldres{\scalebox{0.78}{0.161}} &\boldres{\scalebox{0.78}{0.254}} &\scalebox{0.78}{0.184} &\scalebox{0.78}{0.289} & \scalebox{0.78}{0.199}& \scalebox{0.78}{0.315} &\scalebox{0.78}{0.213} &\scalebox{0.78}{0.316} &\scalebox{0.78}{0.196} &\secondres{\scalebox{0.78}{0.285}} &\scalebox{0.78}{0.201} &\scalebox{0.78}{0.315} &\secondres{\scalebox{0.78}{0.182}} &\scalebox{0.78}{0.286} &\scalebox{0.78}{0.222} &\scalebox{0.78}{0.334} &\scalebox{0.78}{0.378} &\scalebox{0.78}{0.443} &\scalebox{0.78}{0.296} &\scalebox{0.78}{0.386} &\scalebox{0.78}{0.266} &\scalebox{0.78}{0.368}\\
    & \scalebox{0.78}{336} &\boldres{\scalebox{0.78}{0.173}} &\boldres{\scalebox{0.78}{0.267}} &\secondres{\scalebox{0.78}{0.198}} &\secondres{\scalebox{0.78}{0.300}} & \scalebox{0.78}{0.212} &  \scalebox{0.78}{0.329} &\scalebox{0.78}{0.230} &\scalebox{0.78}{0.333} &\scalebox{0.78}{0.209} &\scalebox{0.78}{0.301} &\scalebox{0.78}{0.214} &\scalebox{0.78}{0.329} &\scalebox{0.78}{0.200} &\scalebox{0.78}{0.304} &\scalebox{0.78}{0.231} &\scalebox{0.78}{0.338} &\scalebox{0.78}{0.376} &\scalebox{0.78}{0.443} &\scalebox{0.78}{0.300} &\scalebox{0.78}{0.394} &\scalebox{0.78}{0.280} &\scalebox{0.78}{0.380} \\
    & \scalebox{0.78}{720} &\boldres{\scalebox{0.78}{0.209}} &\boldres{\scalebox{0.78}{0.299}} &\secondres{\scalebox{0.78}{0.220}} &\secondres{\scalebox{0.78}{0.320}} & \scalebox{0.78}{0.233} & \scalebox{0.78}{0.345} &\scalebox{0.78}{0.265} &\scalebox{0.78}{0.360} &\scalebox{0.78}{0.245} &\scalebox{0.78}{0.333} &\scalebox{0.78}{0.246} &\scalebox{0.78}{0.355} &\scalebox{0.78}{0.222} &\scalebox{0.78}{0.321} &\scalebox{0.78}{0.254} &\scalebox{0.78}{0.361} &\scalebox{0.78}{0.376} &\scalebox{0.78}{0.445} &\scalebox{0.78}{0.373} &\scalebox{0.78}{0.439} &\scalebox{0.78}{0.283} &\scalebox{0.78}{0.376} \\
    \cmidrule(lr){2-24}
    & \scalebox{0.78}{Avg} &\boldres{\scalebox{0.78}{0.171}} &\boldres{\scalebox{0.78}{0.246}} &\secondres{\scalebox{0.78}{0.192}} &\secondres{\scalebox{0.78}{0.295}} &\scalebox{0.78}{0.208} &  \scalebox{0.78}{0.323}&\scalebox{0.78}{0.229} &\scalebox{0.78}{0.329} &\scalebox{0.78}{0.212} &\scalebox{0.78}{0.300} &\scalebox{0.78}{0.214} &\scalebox{0.78}{0.327} &\scalebox{0.78}{0.193} &\scalebox{0.78}{0.296} &\scalebox{0.78}{0.227} &\scalebox{0.78}{0.338} &\scalebox{0.78}{0.379} &\scalebox{0.78}{0.445} &\scalebox{0.78}{0.311} &\scalebox{0.78}{0.397} &\scalebox{0.78}{0.272} &\scalebox{0.78}{0.370} \\
    \midrule
    \multirow{5}{*}{\rotatebox{90}{\scalebox{0.95}{Trafﬁc}}} 
    & \scalebox{0.78}{96} &\boldres{\scalebox{0.78}{0.438}} &\boldres{\scalebox{0.78}{0.292}} &\scalebox{0.78}{0.593} &\secondres{\scalebox{0.78}{0.321}} & \scalebox{0.78}{0.607} & \scalebox{0.78}{0.392} &\scalebox{0.78}{0.615} &\scalebox{0.78}{0.391} &\scalebox{0.78}{0.650} &\scalebox{0.78}{0.396} &\secondres{\scalebox{0.78}{0.587}} &\scalebox{0.78}{0.366} &\scalebox{0.78}{0.612} &\scalebox{0.78}{0.338} &\scalebox{0.78}{0.613} &\scalebox{0.78}{0.388} &\scalebox{0.78}{0.867} &\scalebox{0.78}{0.468} &\scalebox{0.78}{0.719} &\scalebox{0.78}{0.391} &\scalebox{0.78}{0.684} &\scalebox{0.78}{0.384} \\
    & \scalebox{0.78}{192} &\boldres{\scalebox{0.78}{0.451}} &\boldres{\scalebox{0.78}{0.298}} &\scalebox{0.78}{0.617} &\secondres{\scalebox{0.78}{0.336}} & \scalebox{0.78}{0.621} & \scalebox{0.78}{0.399} &\scalebox{0.78}{0.601} &\scalebox{0.78}{0.382} &\secondres{\scalebox{0.78}{0.598}} &\scalebox{0.78}{0.370} &\scalebox{0.78}{0.604} &\scalebox{0.78}{0.373} &\scalebox{0.78}{0.613} &\scalebox{0.78}{0.340} &\scalebox{0.78}{0.616} &\scalebox{0.78}{0.382} &\scalebox{0.78}{0.869} &\scalebox{0.78}{0.467} &\scalebox{0.78}{0.696} &\scalebox{0.78}{0.379} &\scalebox{0.78}{0.685} &\scalebox{0.78}{0.390} \\
    & \scalebox{0.78}{336} &\boldres{\scalebox{0.78}{0.472}} &\boldres{\scalebox{0.78}{0.305}} &\scalebox{0.78}{0.629} &\scalebox{0.78}{0.336} & \scalebox{0.78}{0.622}& \scalebox{0.78}{0.396} &\scalebox{0.78}{0.613} &\scalebox{0.78}{0.386} &\secondres{\scalebox{0.78}{0.605}} &\scalebox{0.78}{0.373} &\scalebox{0.78}{0.621} &\scalebox{0.78}{0.383} &\scalebox{0.78}{0.618} &\secondres{\scalebox{0.78}{0.328}} &\scalebox{0.78}{0.622} &\scalebox{0.78}{0.337} &\scalebox{0.78}{0.881} &\scalebox{0.78}{0.469} &\scalebox{0.78}{0.777} &\scalebox{0.78}{0.420} &\scalebox{0.78}{0.734} &\scalebox{0.78}{0.408} \\
    & \scalebox{0.78}{720} &\boldres{\scalebox{0.78}{0.499}} &\boldres{\scalebox{0.78}{0.321}} &\scalebox{0.78}{0.640} &\secondres{\scalebox{0.78}{0.350}} & \scalebox{0.78}{0.632} & \scalebox{0.78}{0.396}  &\scalebox{0.78}{0.658} &\scalebox{0.78}{0.407} &\scalebox{0.78}{0.645} &\scalebox{0.78}{0.394} &\secondres{\scalebox{0.78}{0.626}} &\scalebox{0.78}{0.382} &\scalebox{0.78}{0.653} &\scalebox{0.78}{0.355} &\scalebox{0.78}{0.660} &\scalebox{0.78}{0.408} &\scalebox{0.78}{0.896} &\scalebox{0.78}{0.473} &\scalebox{0.78}{0.864} &\scalebox{0.78}{0.472} &\scalebox{0.78}{0.717} &\scalebox{0.78}{0.396} \\
    \cmidrule(lr){2-24}
    & \scalebox{0.78}{Avg} &\boldres{\scalebox{0.78}{0.465}} &\boldres{\scalebox{0.78}{0.304}} &\scalebox{0.78}{0.620} &\secondres{\scalebox{0.78}{0.336}} & \scalebox{0.78}{0.621} & \scalebox{0.78}{0.396}  &\scalebox{0.78}{0.622} &\scalebox{0.78}{0.392} &\scalebox{0.78}{0.625} &\scalebox{0.78}{0.383} &\secondres{\scalebox{0.78}{0.610}} &\scalebox{0.78}{0.376} &\scalebox{0.78}{0.624} &\scalebox{0.78}{0.340} &\scalebox{0.78}{0.628} &\scalebox{0.78}{0.379} &\scalebox{0.78}{0.878} &\scalebox{0.78}{0.469} &\scalebox{0.78}{0.764} &\scalebox{0.78}{0.416} &\scalebox{0.78}{0.705} &\scalebox{0.78}{0.395} \\
    \midrule
    \multirow{5}{*}{\rotatebox{90}{\scalebox{0.95}{Weather}}} 
    &  \scalebox{0.78}{96} &\boldres{\scalebox{0.78}{0.163}} &\boldres{\scalebox{0.78}{0.207}} &\secondres{\scalebox{0.78}{0.172}} &\secondres{\scalebox{0.78}{0.220}} & \scalebox{0.78}{0.197} & \scalebox{0.78}{0.281} & \scalebox{0.78}{0.182} &\scalebox{0.78}{0.242} & \scalebox{0.78}{0.196} &\scalebox{0.78}{0.255} & \scalebox{0.78}{0.217} &\scalebox{0.78}{0.296} & \scalebox{0.78}{0.173} &\scalebox{0.78}{0.223} & \scalebox{0.78}{0.266} &\scalebox{0.78}{0.336} & \scalebox{0.78}{0.622} &\scalebox{0.78}{0.556} & \scalebox{0.78}{0.300} &\scalebox{0.78}{0.384} & \scalebox{0.78}{0.458} &\scalebox{0.78}{0.490}\\
    & \scalebox{0.78}{192} &\boldres{\scalebox{0.78}{0.214}} &\boldres{\scalebox{0.78}{0.253}} &\secondres{\scalebox{0.78}{0.219}} &\secondres{\scalebox{0.78}{0.261}} & \scalebox{0.78}{0.237} & \scalebox{0.78}{0.312} &\scalebox{0.78}{0.227} &\scalebox{0.78}{0.287} & \scalebox{0.78}{0.237} &\scalebox{0.78}{0.296} & \scalebox{0.78}{0.276} &\scalebox{0.78}{0.336} & \scalebox{0.78}{0.245} &\scalebox{0.78}{0.285} & \scalebox{0.78}{0.307} &\scalebox{0.78}{0.367} & \scalebox{0.78}{0.739} &\scalebox{0.78}{0.624} & \scalebox{0.78}{0.598} &\scalebox{0.78}{0.544} & \scalebox{0.78}{0.658} &\scalebox{0.78}{0.589} \\
    & \scalebox{0.78}{336} &\boldres{\scalebox{0.78}{0.271}} &\boldres{\scalebox{0.78}{0.294}} &\secondres{\scalebox{0.78}{0.280}} &\secondres{\scalebox{0.78}{0.306}} & \scalebox{0.78}{0.298} & \scalebox{0.78}{0.353} &\scalebox{0.78}{0.282} &\scalebox{0.78}{0.334} & \scalebox{0.78}{0.283} &\scalebox{0.78}{0.335} & \scalebox{0.78}{0.339} &\scalebox{0.78}{0.380} & \scalebox{0.78}{0.321} &\scalebox{0.78}{0.338} & \scalebox{0.78}{0.359} &\scalebox{0.78}{0.395} & \scalebox{0.78}{1.004} &\scalebox{0.78}{0.753} & \scalebox{0.78}{0.578} &\scalebox{0.78}{0.523} & \scalebox{0.78}{0.797} &\scalebox{0.78}{0.652} \\
    & \scalebox{0.78}{720} &\scalebox{0.78}{0.360} &\boldres{\scalebox{0.78}{0.346}} &\scalebox{0.78}{0.365} &\secondres{\scalebox{0.78}{0.359}} & \secondres{\scalebox{0.78}{0.352}} & \scalebox{0.78}{0.288} &\secondres{\scalebox{0.78}{0.352}} &\scalebox{0.78}{0.386} & \boldres{\scalebox{0.78}{0.345}} &\scalebox{0.78}{0.381} & \scalebox{0.78}{0.403} &\scalebox{0.78}{0.428} & \scalebox{0.78}{0.414} &\scalebox{0.78}{0.410} & \scalebox{0.78}{0.419} &\scalebox{0.78}{0.428} & \scalebox{0.78}{1.420} &\scalebox{0.78}{0.934} & \scalebox{0.78}{1.059} &\scalebox{0.78}{0.741} & \scalebox{0.78}{0.869} &\scalebox{0.78}{0.675} \\
    \cmidrule(lr){2-24}
    & \scalebox{0.78}{Avg} &\boldres{\scalebox{0.78}{0.249}} &\boldres{\scalebox{0.78}{0.275}} &\secondres{\scalebox{0.78}{0.259}} &\secondres{\scalebox{0.78}{0.287}} & \scalebox{0.78}{0.271} &  \scalebox{0.78}{0.334}& \scalebox{0.78}{0.261} &\scalebox{0.78}{0.312} &\scalebox{0.78}{0.265} &\scalebox{0.78}{0.317} &\scalebox{0.78}{0.309} &\scalebox{0.78}{0.360} &\scalebox{0.78}{0.288} &\scalebox{0.78}{0.314} &\scalebox{0.78}{0.338} &\scalebox{0.78}{0.382} &\scalebox{0.78}{0.946} &\scalebox{0.78}{0.717} &\scalebox{0.78}{0.634} &\scalebox{0.78}{0.548} &\scalebox{0.78}{0.696} &\scalebox{0.78}{0.602} \\
    \midrule
    \multirow{5}{*}{\rotatebox{90}{\scalebox{0.95}{ETTh1}}}
    &  \scalebox{0.78}{96} &\scalebox{0.78}{0.392} &\scalebox{0.78}{0.409} &\secondres{\scalebox{0.78}{0.384}} &\secondres{\scalebox{0.78}{0.402}} & \scalebox{0.78}{0.494} & \scalebox{0.78}{0.479} &\scalebox{0.78}{0.424} &\scalebox{0.78}{0.432} & \scalebox{0.78}{0.386} &\boldres{\scalebox{0.78}{0.400}} &\boldres{\scalebox{0.78}{0.376}} &\scalebox{0.78}{0.419} &\scalebox{0.78}{0.513} &\scalebox{0.78}{0.491} &\scalebox{0.78}{0.449} &\scalebox{0.78}{0.459} &\scalebox{0.78}{0.664} &\scalebox{0.78}{0.612} &\scalebox{0.78}{0.865} &\scalebox{0.78}{0.713} &\scalebox{0.78}{0.878} &\scalebox{0.78}{0.740} \\
    & \scalebox{0.78}{192} &\scalebox{0.78}{0.445} &\scalebox{0.78}{0.436} &\secondres{\scalebox{0.78}{0.436}} &\boldres{\scalebox{0.78}{0.429}} & \scalebox{0.78}{0.538}& \scalebox{0.78}{0.504} &\scalebox{0.78}{0.475} &\scalebox{0.78}{0.462} &\scalebox{0.78}{0.437} &\secondres{\scalebox{0.78}{0.432}} &\boldres{\scalebox{0.78}{0.420}} &\scalebox{0.78}{0.448} &\scalebox{0.78}{0.534} &\scalebox{0.78}{0.504} &\scalebox{0.78}{0.500} &\scalebox{0.78}{0.482} &\scalebox{0.78}{0.790} &\scalebox{0.78}{0.681} &\scalebox{0.78}{1.008} &\scalebox{0.78}{0.792} &\scalebox{0.78}{1.037} &\scalebox{0.78}{0.824} \\
    & \scalebox{0.78}{336} &\scalebox{0.78}{0.482} &\boldres{\scalebox{0.78}{0.456}} &\scalebox{0.78}{0.491} &\scalebox{0.78}{0.469}  & \scalebox{0.78}{0.574}& \scalebox{0.78}{0.521} &\scalebox{0.78}{0.518} &\scalebox{0.78}{0.488} &\secondres{\scalebox{0.78}{0.481}} & \secondres{\scalebox{0.78}{0.459}} &\boldres{\scalebox{0.78}{0.459}} &\scalebox{0.78}{0.465} &\scalebox{0.78}{0.588} &\scalebox{0.78}{0.535} &\scalebox{0.78}{0.521} &\scalebox{0.78}{0.496} &\scalebox{0.78}{0.891} &\scalebox{0.78}{0.738} &\scalebox{0.78}{1.107} &\scalebox{0.78}{0.809} &\scalebox{0.78}{1.238} &\scalebox{0.78}{0.932} \\
    & \scalebox{0.78}{720} &\boldres{\scalebox{0.78}{0.489}} &\boldres{\scalebox{0.78}{0.480}} &\scalebox{0.78}{0.521} &\secondres{\scalebox{0.78}{0.500}}  & \scalebox{0.78}{0.562}&  \scalebox{0.78}{0.535}& \scalebox{0.78}{0.547} &\scalebox{0.78}{0.533} &\scalebox{0.78}{0.519} &\scalebox{0.78}{0.516} &\secondres{\scalebox{0.78}{0.506}} &\scalebox{0.78}{0.507} &\scalebox{0.78}{0.643} &\scalebox{0.78}{0.616} &\scalebox{0.78}{0.514} &\scalebox{0.78}{0.512} &\scalebox{0.78}{0.963} &\scalebox{0.78}{0.782} &\scalebox{0.78}{1.181} &\scalebox{0.78}{0.865} &\scalebox{0.78}{1.135} &\scalebox{0.78}{0.852} \\
    \cmidrule(lr){2-24}
    & \scalebox{0.78}{Avg} &\secondres{\scalebox{0.78}{0.452}} &\boldres{\scalebox{0.78}{0.445}} &\scalebox{0.78}{0.458} &\secondres{\scalebox{0.78}{0.450}} & \scalebox{0.78}{0.542} &  \scalebox{0.78}{0.510}& \scalebox{0.78}{0.491} &\scalebox{0.78}{0.479} & \scalebox{0.78}{0.456} &\scalebox{0.78}{0.452} &\boldres{\scalebox{0.78}{0.440}} &\scalebox{0.78}{0.460} &\scalebox{0.78}{0.570} &\scalebox{0.78}{0.537} &\scalebox{0.78}{0.496} &\scalebox{0.78}{0.487} &\scalebox{0.78}{0.827} &\scalebox{0.78}{0.703} &\scalebox{0.78}{1.040} &\scalebox{0.78}{0.795} &\scalebox{0.78}{1.072} &\scalebox{0.78}{0.837} \\
    \midrule
    \multirow{5}{*}{\rotatebox{90}{\scalebox{0.95}{ETTh2}}}  
    &  \scalebox{0.78}{96} & \boldres{\scalebox{0.78}{0.305}} & \boldres{\scalebox{0.78}{0.353}} & \scalebox{0.78}{0.340} & \scalebox{0.78}{0.374} & \scalebox{0.78}{0.340}& \scalebox{0.78}{0.391} &\scalebox{0.78}{0.397} &\scalebox{0.78}{0.437} &\secondres{\scalebox{0.78}{0.333}} &\secondres{\scalebox{0.78}{0.387}}  &\scalebox{0.78}{0.358} &\scalebox{0.78}{0.397} &\scalebox{0.78}{0.476} &\scalebox{0.78}{0.458} &\scalebox{0.78}{0.346} &\scalebox{0.78}{0.388} &\scalebox{0.78}{0.645} &\scalebox{0.78}{0.597} &\scalebox{0.78}{3.755} &\scalebox{0.78}{1.525} &\scalebox{0.78}{2.116} &\scalebox{0.78}{1.197} \\
    & \scalebox{0.78}{192} & \boldres{\scalebox{0.78}{0.382}} & \boldres{\scalebox{0.78}{0.401}} & \secondres{\scalebox{0.78}{0.402}} & \secondres{\scalebox{0.78}{0.414}} & \scalebox{0.78}{0.430} & \scalebox{0.78}{0.439} &\scalebox{0.78}{0.520} &\scalebox{0.78}{0.504} &\scalebox{0.78}{0.477} &\scalebox{0.78}{0.476} &\scalebox{0.78}{0.429} &\scalebox{0.78}{0.439} &\scalebox{0.78}{0.512} &\scalebox{0.78}{0.493} &\scalebox{0.78}{0.456} &\scalebox{0.78}{0.452}  &\scalebox{0.78}{0.788} &\scalebox{0.78}{0.683} &\scalebox{0.78}{5.602} &\scalebox{0.78}{1.931} &\scalebox{0.78}{4.315} &\scalebox{0.78}{1.635} \\
    & \scalebox{0.78}{336} & \boldres{\scalebox{0.78}{0.434}} & \boldres{\scalebox{0.78}{0.445}} & \secondres{\scalebox{0.78}{0.452}} & \secondres{\scalebox{0.78}{0.452}} & \scalebox{0.78}{0.485} &  \scalebox{0.78}{0.479} &\scalebox{0.78}{0.626} &\scalebox{0.78}{0.559} &\scalebox{0.78}{0.594} &\scalebox{0.78}{0.541} &\scalebox{0.78}{0.496} &\scalebox{0.78}{0.487} &\scalebox{0.78}{0.552} &\scalebox{0.78}{0.551} &\scalebox{0.78}{0.482} &\scalebox{0.78}{0.486}  &\scalebox{0.78}{0.907} &\scalebox{0.78}{0.747} &\scalebox{0.78}{4.721} &\scalebox{0.78}{1.835} &\scalebox{0.78}{1.124} &\scalebox{0.78}{1.604} \\
    & \scalebox{0.78}{720} & \boldres{\scalebox{0.78}{0.424}} & \boldres{\scalebox{0.78}{0.444}} & \secondres{\scalebox{0.78}{0.462}} & \secondres{\scalebox{0.78}{0.468}} & \scalebox{0.78}{0.500} & \scalebox{0.78}{0.497}  &\scalebox{0.78}{0.863} &\scalebox{0.78}{0.672} &\scalebox{0.78}{0.831} &\scalebox{0.78}{0.657} &\scalebox{0.78}{0.463} &\scalebox{0.78}{0.474} &\scalebox{0.78}{0.562} &\scalebox{0.78}{0.560} &\scalebox{0.78}{0.515} &\scalebox{0.78}{0.511}  &\scalebox{0.78}{0.963} &\scalebox{0.78}{0.783} &\scalebox{0.78}{3.647} &\scalebox{0.78}{1.625} &\scalebox{0.78}{3.188} &\scalebox{0.78}{1.540} \\
    \cmidrule(lr){2-24}
    & \scalebox{0.78}{Avg} &\boldres{\scalebox{0.78}{0.386}} &\boldres{\scalebox{0.78}{0.411}} &\secondres{\scalebox{0.78}{0.414}} &\secondres{\scalebox{0.78}{0.427}} & \scalebox{0.78}{0.439}& \scalebox{0.78}{0.452} &\scalebox{0.78}{0.602} &\scalebox{0.78}{0.543} &\scalebox{0.78}{0.559} &\scalebox{0.78}{0.515} &\scalebox{0.78}{0.437} &\scalebox{0.78}{0.449} &\scalebox{0.78}{0.526} &\scalebox{0.78}{0.516} &\scalebox{0.78}{0.450} &\scalebox{0.78}{0.459} &\scalebox{0.78}{0.826} &\scalebox{0.78}{0.703} &\scalebox{0.78}{4.431} &\scalebox{0.78}{1.729} &\scalebox{0.78}{2.686} &\scalebox{0.78}{1.494} \\
    \midrule
    % continue here
    \multirow{5}{*}{\rotatebox{90}{\scalebox{0.95}{ETTm1}}}
    &  \scalebox{0.78}{96} &\boldres{\scalebox{0.78}{0.336}} &\boldres{\scalebox{0.78}{0.369}} &\secondres{\scalebox{0.78}{0.338}} &\scalebox{0.78}{0.375} & \scalebox{0.78}{0.375} & \scalebox{0.78}{0.398} &\scalebox{0.78}{0.374} &\scalebox{0.78}{0.400} &\scalebox{0.78}{0.345} &\secondres{\scalebox{0.78}{0.372}}  &\scalebox{0.78}{0.379} &\scalebox{0.78}{0.419} &\scalebox{0.78}{0.386} &\scalebox{0.78}{0.398} &\scalebox{0.78}{0.505} &\scalebox{0.78}{0.475} &\scalebox{0.78}{0.543} &\scalebox{0.78}{0.510} &\scalebox{0.78}{0.672} &\scalebox{0.78}{0.571} &\scalebox{0.78}{0.600} &\scalebox{0.78}{0.546} \\
    & \scalebox{0.78}{192} &\boldres{\scalebox{0.78}{0.374}} &\boldres{\scalebox{0.78}{0.387}}  &\boldres{\scalebox{0.78}{0.374}} &\boldres{\scalebox{0.78}{0.387}} & \scalebox{0.78}{0.408}&\scalebox{0.78}{0.410} &\scalebox{0.78}{0.400} &\scalebox{0.78}{0.407} &\secondres{\scalebox{0.78}{0.380}} &\secondres{\scalebox{0.78}{0.389}} &\scalebox{0.78}{0.426} &\scalebox{0.78}{0.441} &\scalebox{0.78}{0.459} &\scalebox{0.78}{0.444} &\scalebox{0.78}{0.553} &\scalebox{0.78}{0.496} &\scalebox{0.78}{0.557} &\scalebox{0.78}{0.537} &\scalebox{0.78}{0.795} &\scalebox{0.78}{0.669} &\scalebox{0.78}{0.837} &\scalebox{0.78}{0.700} \\
    & \scalebox{0.78}{336} &\boldres{\scalebox{0.78}{0.408}} &\boldres{\scalebox{0.78}{0.407}} &\secondres{\scalebox{0.78}{0.410}} &\secondres{\scalebox{0.78}{0.411}}  &\scalebox{0.78}{0.435} & \scalebox{0.78}{0.428} &\scalebox{0.78}{0.438} &\scalebox{0.78}{0.438} &\scalebox{0.78}{0.413} &\scalebox{0.78}{0.413} &\scalebox{0.78}{0.445} &\scalebox{0.78}{0.459} &\scalebox{0.78}{0.495} &\scalebox{0.78}{0.464} &\scalebox{0.78}{0.621} &\scalebox{0.78}{0.537} &\scalebox{0.78}{0.754} &\scalebox{0.78}{0.655} &\scalebox{0.78}{1.212} &\scalebox{0.78}{0.871} &\scalebox{0.78}{1.124} &\scalebox{0.78}{0.832}\\
    & \scalebox{0.78}{720} &\secondres{\scalebox{0.78}{0.477}} &\boldres{\scalebox{0.78}{0.442}} &\scalebox{0.78}{0.478} &\secondres{\scalebox{0.78}{0.450}}  &\scalebox{0.78}{0.499} &\scalebox{0.78}{0.462} &\scalebox{0.78}{0.527} &\scalebox{0.78}{0.502} &\secondres{\scalebox{0.78}{0.474}} &\scalebox{0.78}{0.453} &\scalebox{0.78}{0.543} &\scalebox{0.78}{0.490} &\scalebox{0.78}{0.585} &\scalebox{0.78}{0.516} &\scalebox{0.78}{0.671} &\scalebox{0.78}{0.561} &\scalebox{0.78}{0.908} &\scalebox{0.78}{0.724} &\scalebox{0.78}{1.166} &\scalebox{0.78}{0.823} &\scalebox{0.78}{1.153} &\scalebox{0.78}{0.820} \\
    \cmidrule(lr){2-24}
    & \scalebox{0.78}{Avg} &\boldres{\scalebox{0.78}{0.399}} &\boldres{\scalebox{0.78}{0.401}} &\secondres{\scalebox{0.78}{0.400}} &\secondres{\scalebox{0.78}{0.406}}  &\scalebox{0.78}{0.429} & \scalebox{0.78}{0.425}&\scalebox{0.78}{0.435} &\scalebox{0.78}{0.437} &\scalebox{0.78}{0.403} &\scalebox{0.78}{0.407} &\scalebox{0.78}{0.448} &\scalebox{0.78}{0.452} &\scalebox{0.78}{0.481} &\scalebox{0.78}{0.456} &\scalebox{0.78}{0.588} &\scalebox{0.78}{0.517} &\scalebox{0.78}{0.691} &\scalebox{0.78}{0.607} &\scalebox{0.78}{0.961} &\scalebox{0.78}{0.734} &\scalebox{0.78}{0.929} &\scalebox{0.78}{0.725} \\
    \midrule
    \multirow{5}{*}{\rotatebox{90}{\scalebox{0.95}{ETTm2}}}
    &  \scalebox{0.78}{96} &\boldres{\scalebox{0.78}{0.184}} &\boldres{\scalebox{0.78}{0.267}} &\secondres{\scalebox{0.78}{0.187}} &\boldres{\scalebox{0.78}{0.267}} & \scalebox{0.78}{0.189} & \scalebox{0.78}{0.280}& \scalebox{0.78}{0.209} &\scalebox{0.78}{0.308} &\scalebox{0.78}{0.193} &\scalebox{0.78}{0.292} &\scalebox{0.78}{0.203} &\scalebox{0.78}{0.287} &\scalebox{0.78}{0.192} &\secondres{\scalebox{0.78}{0.274}} &\scalebox{0.78}{0.255} &\scalebox{0.78}{0.339} &\scalebox{0.78}{0.435} &\scalebox{0.78}{0.507} &\scalebox{0.78}{0.365} &\scalebox{0.78}{0.453} &\scalebox{0.78}{0.768} &\scalebox{0.78}{0.642} \\
    & \scalebox{0.78}{192} &\secondres{\scalebox{0.78}{0.252}} &\boldres{\scalebox{0.78}{0.307}} &\boldres{\scalebox{0.78}{0.249}} &\secondres{\scalebox{0.78}{0.309}} & \scalebox{0.78}{0.253} & \scalebox{0.78}{0.319} & \scalebox{0.78}{0.311} &\scalebox{0.78}{0.382} &\scalebox{0.78}{0.284} &\scalebox{0.78}{0.362} &\scalebox{0.78}{0.269} &\scalebox{0.78}{0.328} &\scalebox{0.78}{0.280} &\scalebox{0.78}{0.339} &\scalebox{0.78}{0.281} &\scalebox{0.78}{0.340} &\scalebox{0.78}{0.730} &\scalebox{0.78}{0.673} &\scalebox{0.78}{0.533} &\scalebox{0.78}{0.563} &\scalebox{0.78}{0.989} &\scalebox{0.78}{0.757} \\
    & \scalebox{0.78}{336} &\boldres{\scalebox{0.78}{0.314}} &\boldres{\scalebox{0.78}{0.345}} &\secondres{\scalebox{0.78}{0.321}} &\secondres{\scalebox{0.78}{0.351}}  & \boldres{\scalebox{0.78}{0.314}}&\scalebox{0.78}{0.357} &\scalebox{0.78}{0.442} &\scalebox{0.78}{0.466} &\scalebox{0.78}{0.369} &\scalebox{0.78}{0.427} &\scalebox{0.78}{0.325} &\scalebox{0.78}{0.366} &\scalebox{0.78}{0.334} &\scalebox{0.78}{0.361} &\scalebox{0.78}{0.339} &\scalebox{0.78}{0.372} &\scalebox{0.78}{1.201} &\scalebox{0.78}{0.845} &\scalebox{0.78}{1.363} &\scalebox{0.78}{0.887} &\scalebox{0.78}{1.334} &\scalebox{0.78}{0.872} \\
    & \scalebox{0.78}{720} &\secondres{\scalebox{0.78}{0.412}} &\boldres{\scalebox{0.78}{0.402}} &\boldres{\scalebox{0.78}{0.408}} &\secondres{\scalebox{0.78}{0.403}} & \scalebox{0.78}{0.414} & \scalebox{0.78}{0.413}& \scalebox{0.78}{0.675} &\scalebox{0.78}{0.587} &\scalebox{0.78}{0.554} &\scalebox{0.78}{0.522} &\scalebox{0.78}{0.421} &\scalebox{0.78}{0.415} &\scalebox{0.78}{0.417} &\scalebox{0.78}{0.413} &\scalebox{0.78}{0.433} &\scalebox{0.78}{0.432} &\scalebox{0.78}{3.625} &\scalebox{0.78}{1.451} &\scalebox{0.78}{3.379} &\scalebox{0.78}{1.338} &\scalebox{0.78}{3.048} &\scalebox{0.78}{1.328} \\
    \cmidrule(lr){2-24}
    & \scalebox{0.78}{Avg} &\boldres{\scalebox{0.78}{0.291}} &\boldres{\scalebox{0.78}{0.330}} &\boldres{\scalebox{0.78}{0.291}} &\secondres{\scalebox{0.78}{0.333}} & \secondres{\scalebox{0.78}{0.293}} & \scalebox{0.78}{0.342} & \scalebox{0.78}{0.409} &\scalebox{0.78}{0.436} &\scalebox{0.78}{0.350} &\scalebox{0.78}{0.401} &\scalebox{0.78}{0.305} &\scalebox{0.78}{0.349} &\scalebox{0.78}{0.306} &\scalebox{0.78}{0.347} &\scalebox{0.78}{0.327} &\scalebox{0.78}{0.371} &\scalebox{0.78}{1.498} &\scalebox{0.78}{0.869} &\scalebox{0.78}{1.410} &\scalebox{0.78}{0.810} &\scalebox{0.78}{1.535} &\scalebox{0.78}{0.900} \\
    \midrule
    
    \multirow{5}{*}{\rotatebox{90}{\scalebox{0.95}{Exchange}}} 
    &  \scalebox{0.78}{96} &\secondres{\scalebox{0.78}{0.086}} &\secondres{\scalebox{0.78}{0.206}} &\scalebox{0.78}{0.107} &\scalebox{0.78}{0.234} & \boldres{\scalebox{0.78}{0.085}} & \boldres{\scalebox{0.78}{0.204}}  &\scalebox{0.78}{0.116} &\scalebox{0.78}{0.262} &\scalebox{0.78}{0.088} &\scalebox{0.78}{0.218} &\scalebox{0.78}{0.148} &\scalebox{0.78}{0.278} &\scalebox{0.78}{0.111} &\scalebox{0.78}{0.237} &\scalebox{0.78}{0.197} &\scalebox{0.78}{0.323} &\scalebox{0.78}{1.748} &\scalebox{0.78}{1.105} &\scalebox{0.78}{0.847} &\scalebox{0.78}{0.752} &\scalebox{0.78}{0.968} &\scalebox{0.78}{0.812} \\
    & \scalebox{0.78}{192}  &\boldres{\scalebox{0.78}{0.176}} &\boldres{\scalebox{0.78}{0.299}} &\scalebox{0.78}{0.226} &\scalebox{0.78}{0.344} & \secondres{\scalebox{0.78}{0.182}} & \secondres{\scalebox{0.78}{0.303}} &  \scalebox{0.78}{0.215} &\scalebox{0.78}{0.359} &\boldres{\scalebox{0.78}{0.176}} &\scalebox{0.78}{0.315} &\scalebox{0.78}{0.271} &\scalebox{0.78}{0.380} &\scalebox{0.78}{0.219} &\scalebox{0.78}{0.335} &\scalebox{0.78}{0.300} &\scalebox{0.78}{0.369} &\scalebox{0.78}{1.874} &\scalebox{0.78}{1.151} &\scalebox{0.78}{1.204} &\scalebox{0.78}{0.895} &\scalebox{0.78}{1.040} &\scalebox{0.78}{0.851} \\
    & \scalebox{0.78}{336}  &\secondres{\scalebox{0.78}{0.330}} & \boldres{\scalebox{0.78}{0.416}} &\scalebox{0.78}{0.367} & \scalebox{0.78}{0.448} & \scalebox{0.78}{0.348} & \scalebox{0.78}{0.428}  &\scalebox{0.78}{0.377} &\scalebox{0.78}{0.466} &\boldres{\scalebox{0.78}{0.313}} &\secondres{\scalebox{0.78}{0.427}} &\scalebox{0.78}{0.460} &\scalebox{0.78}{0.500} &\scalebox{0.78}{0.421} &\scalebox{0.78}{0.476} &\scalebox{0.78}{0.509} &\scalebox{0.78}{0.524} &\scalebox{0.78}{1.943} &\scalebox{0.78}{1.172} &\scalebox{0.78}{1.672} &\scalebox{0.78}{1.036} &\scalebox{0.78}{1.659} &\scalebox{0.78}{1.081} \\
    & \scalebox{0.78}{720}  &\boldres{\scalebox{0.78}{0.828}} &\boldres{\scalebox{0.78}{0.689}} &\scalebox{0.78}{0.964} &\scalebox{0.78}{0.746} & \scalebox{0.78}{1.025} & \scalebox{0.78}{0.774} &\secondres{\scalebox{0.78}{0.831}} &\scalebox{0.78}{0.699} &\scalebox{0.78}{0.839} &\secondres{\scalebox{0.78}{0.695}} &\scalebox{0.78}{1.195} &\scalebox{0.78}{0.841} &\scalebox{0.78}{1.092} &\scalebox{0.78}{0.769} &\scalebox{0.78}{1.447} &\scalebox{0.78}{0.941} &\scalebox{0.78}{2.085} &\scalebox{0.78}{1.206} &\scalebox{0.78}{2.478} &\scalebox{0.78}{1.310} &\scalebox{0.78}{1.941} &\scalebox{0.78}{1.127} \\
    \cmidrule(lr){2-24}
    & \scalebox{0.78}{Avg} &\secondres{\scalebox{0.78}{0.355}} &\boldres{\scalebox{0.78}{0.403}} &\scalebox{0.78}{0.416} &\scalebox{0.78}{0.443} & \scalebox{0.78}{0.410}& \scalebox{0.78}{0.427}  &\scalebox{0.78}{0.385} &\scalebox{0.78}{0.447} &\boldres{\scalebox{0.78}{0.354}} &\secondres{\scalebox{0.78}{0.414}} &\scalebox{0.78}{0.519} &\scalebox{0.78}{0.500} &\scalebox{0.78}{0.461} &\scalebox{0.78}{0.454} &\scalebox{0.78}{0.613} &\scalebox{0.78}{0.539} &\scalebox{0.78}{1.913} &\scalebox{0.78}{1.159} &\scalebox{0.78}{1.550} &\scalebox{0.78}{0.998} &\scalebox{0.78}{1.402} &\scalebox{0.78}{0.968} \\
    \midrule
    \multirow{5}{*}{\rotatebox{90}{\scalebox{0.95}{ILI}}} 
    &  \scalebox{0.78}{24} &\boldres{\scalebox{0.78}{2.033}} &\boldres{\scalebox{0.78}{0.870}} &\scalebox{0.78}{2.317} &\secondres{\scalebox{0.78}{0.934}} & \scalebox{0.78}{2.527} & \scalebox{0.78}{1.020}  &\scalebox{0.78}{8.313} &\scalebox{0.78}{2.144} & \scalebox{0.78}{2.398} &\scalebox{0.78}{1.040} & \scalebox{0.78}{3.228} &\scalebox{0.78}{1.260} & \secondres{\scalebox{0.78}{2.294}} &\scalebox{0.78}{0.945} & \scalebox{0.78}{3.483} &\scalebox{0.78}{1.287} & \scalebox{0.78}{7.394} &\scalebox{0.78}{2.012} & \scalebox{0.78}{5.764} &\scalebox{0.78}{1.677} & \scalebox{0.78}{4.480} &\scalebox{0.78}{1.444} \\
    & \scalebox{0.78}{36}  &\secondres{\scalebox{0.78}{1.909}} &\secondres{\scalebox{0.78}{0.868}} &\scalebox{0.78}{1.972} &\scalebox{0.78}{0.920} & \scalebox{0.78}{2.615} & \scalebox{0.78}{1.007} &\scalebox{0.78}{6.631} &\scalebox{0.78}{1.902} & \scalebox{0.78}{2.646} &\scalebox{0.78}{1.088} & \scalebox{0.78}{2.679} &\scalebox{0.78}{1.080} & \boldres{\scalebox{0.78}{1.825}} &\boldres{\scalebox{0.78}{0.848}} & \scalebox{0.78}{3.103} &\scalebox{0.78}{1.148} & \scalebox{0.78}{7.551} &\scalebox{0.78}{2.031} & \scalebox{0.78}{4.755} &\scalebox{0.78}{1.467} & \scalebox{0.78}{4.799} &\scalebox{0.78}{1.467} \\
    & \scalebox{0.78}{48} &\secondres{\scalebox{0.78}{2.126}} &\secondres{\scalebox{0.78}{0.929}} &\scalebox{0.78}{2.238} &\scalebox{0.78}{0.940} & \scalebox{0.78}{2.359} & \scalebox{0.78}{0.972} &\scalebox{0.78}{7.299} &\scalebox{0.78}{1.982} & \scalebox{0.78}{2.614} &\scalebox{0.78}{1.086} & \scalebox{0.78}{2.622} &\scalebox{0.78}{1.078} & \boldres{\scalebox{0.78}{2.010}} &\boldres{\scalebox{0.78}{0.900}} & \scalebox{0.78}{2.669} &\scalebox{0.78}{1.085} & \scalebox{0.78}{7.662} &\scalebox{0.78}{2.057} & \scalebox{0.78}{4.763} &\scalebox{0.78}{1.469} & \scalebox{0.78}{4.800} &\scalebox{0.78}{1.468} \\
    & \scalebox{0.78}{60} &\secondres{\scalebox{0.78}{2.039}} &\boldres{\scalebox{0.78}{0.914}} &\boldres{\scalebox{0.78}{2.027}} &\secondres{\scalebox{0.78}{0.928}} & \scalebox{0.78}{2.487}&  \scalebox{0.78}{1.016} &\scalebox{0.78}{7.283} &\scalebox{0.78}{1.985} & \scalebox{0.78}{2.804} &\scalebox{0.78}{1.146} & \scalebox{0.78}{2.857} &\scalebox{0.78}{1.157} & \scalebox{0.78}{2.178} &\scalebox{0.78}{0.963} & \scalebox{0.78}{2.770} &\scalebox{0.78}{1.125} & \scalebox{0.78}{7.931} &\scalebox{0.78}{2.100} & \scalebox{0.78}{5.264} &\scalebox{0.78}{1.564} & \scalebox{0.78}{5.278} &\scalebox{0.78}{1.560}  \\
    \cmidrule(lr){2-24}
    & \scalebox{0.78}{Avg} &\boldres{\scalebox{0.78}{2.027}} &\boldres{\scalebox{0.78}{0.895}} &\scalebox{0.78}{2.139} &\scalebox{0.78}{0.931} & \scalebox{0.78}{2.497}& \scalebox{0.78}{1.004}  &\scalebox{0.78}{7.382} &\scalebox{0.78}{2.003} &\scalebox{0.78}{2.616} &\scalebox{0.78}{1.090} &\scalebox{0.78}{2.847} &\scalebox{0.78}{1.144} &\secondres{\scalebox{0.78}{2.077}} &\secondres{\scalebox{0.78}{0.914}} &\scalebox{0.78}{3.006} &\scalebox{0.78}{1.161} &\scalebox{0.78}{7.635} &\scalebox{0.78}{2.050} &\scalebox{0.78}{5.137} &\scalebox{0.78}{1.544} &\scalebox{0.78}{4.839} &\scalebox{0.78}{1.485} \\
    \midrule
    \multicolumn{2}{c}{\scalebox{0.78}{{$1^{\text{st}}$ Count}}} & \multicolumn{2}{c}{\boldres{\scalebox{0.78}{55}}} & \multicolumn{2}{c}{\secondres{\scalebox{0.78}{7}}} & \multicolumn{2}{c}{\scalebox{0.78}{3}} & \multicolumn{2}{c}{\scalebox{0.78}{0}} & \multicolumn{2}{c}{\scalebox{0.78}{4}} & \multicolumn{2}{c}{\scalebox{0.78}{3}} & \multicolumn{2}{c}{\scalebox{0.78}{4}} &  \multicolumn{2}{c}{\scalebox{0.78}{0}} &  \multicolumn{2}{c}{\scalebox{0.78}{0}} & \multicolumn{2}{c}{\scalebox{0.78}{0}} & \multicolumn{2}{c}{\scalebox{0.78}{0}} \\
    \bottomrule
  \end{tabular}
    \end{small}
  \end{threeparttable}
    }  
\end{table}
\section{Correlation coefficient matrix of different datasets} \label{app:corr}
\begin{figure}[htbp]
	\centering
	\begin{minipage}{0.48\linewidth}
		\centering
		\includegraphics[width=1.03\linewidth]{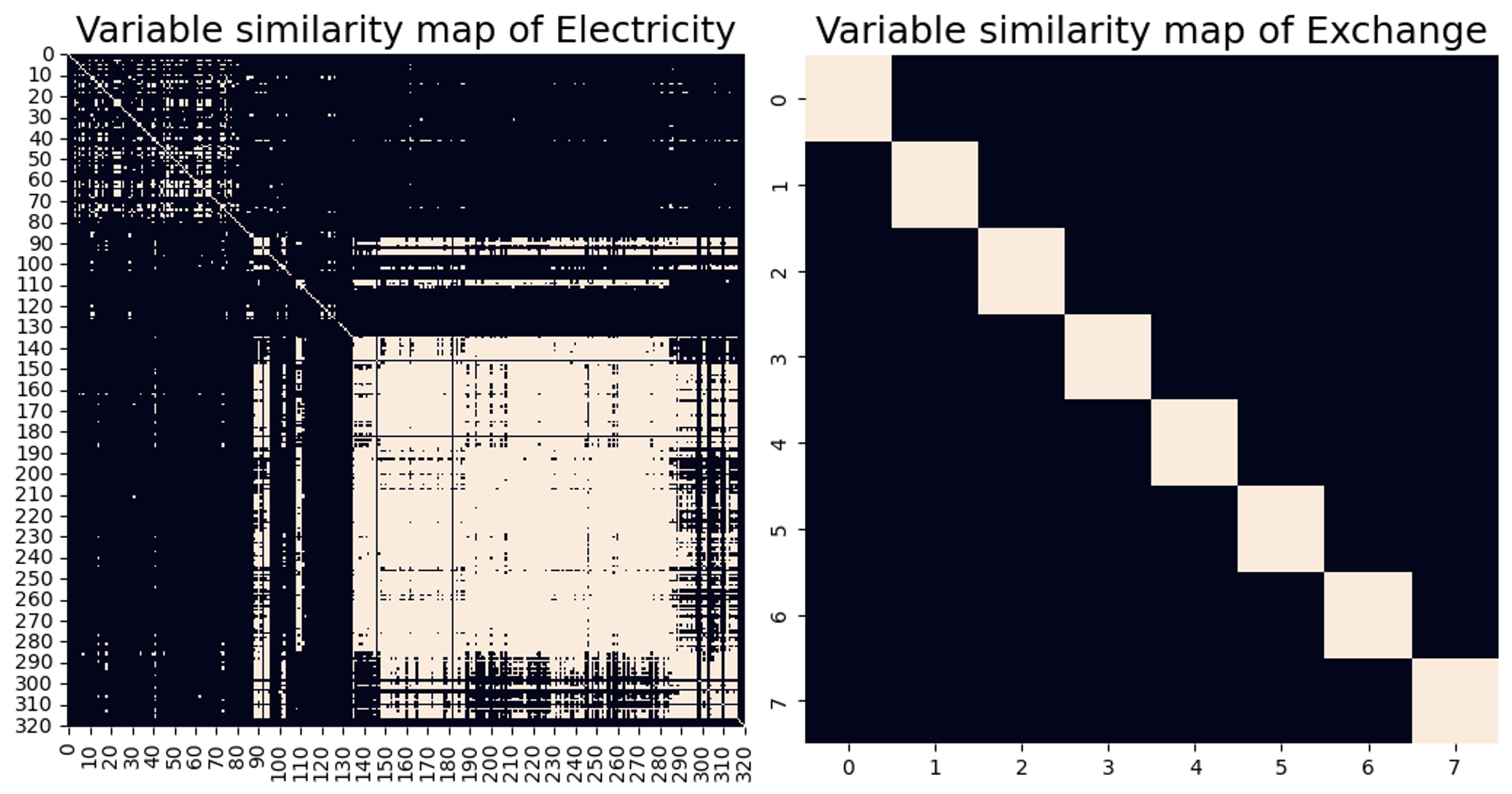}
		\caption{Similarity of different variables of multivariate series. It can be seen that the different variables of Electricity have strong correlation, while the different variables of Exchange have weak correlation.}
		\label{fig:similar_meanwhile}
	\end{minipage}
 \hspace{5pt}
        \begin{minipage}{0.48\linewidth}
		\centering
		\includegraphics[width=1.03\linewidth]{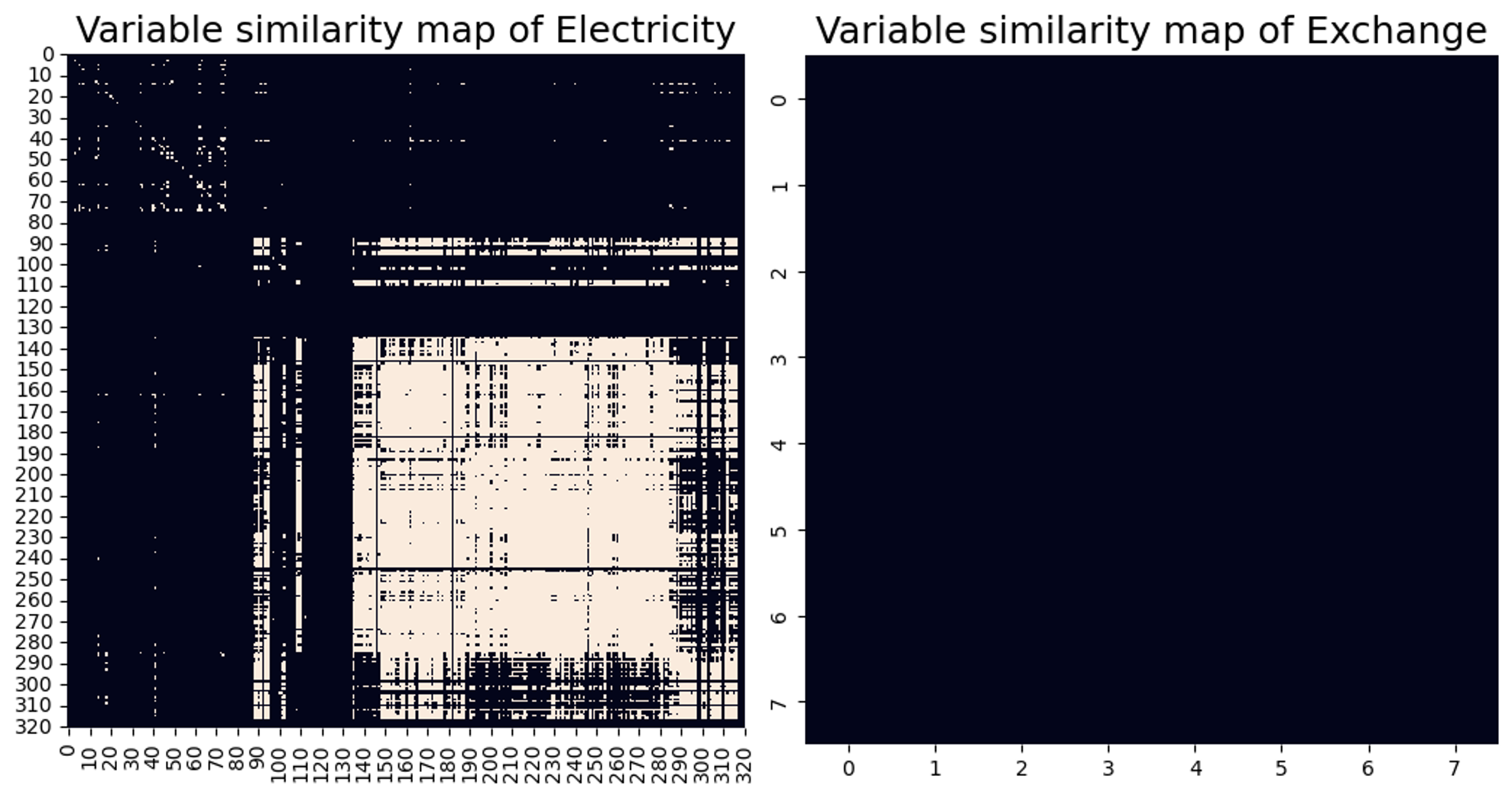}
		\caption{Similarity of historical and future variables of multivariate series. The historical and future variables of Electricity have strong correlation, while the historical and future variables of Exchange have little correlation.}
		\label{fig:similar_different}
	\end{minipage}
\end{figure}

For a more in-depth analysis of the Client's performance on different datasets, we calculate the correlation coefficient matrix of different variables in the Electricity and Exchange datasets. We randomly select 100 sub-series of length 96 (same as the look-back window size) from the series and then calculate the average of their correlation matrices. To strengthen the comparison, we consider two variables to have a strong dependency if their correlation coefficient is greater than 0.8 and set the corresponding value to 1; otherwise, the value is set to 0. Figure \ref{fig:similar_meanwhile} shows the resulting correlation matrices. As we can see, there are strong correlations between different variables in the Electricity dataset, whereas the Exchange dataset has weak correlations between variables. This explains why the Client performs well on the Electricity dataset but mediocrely on the Exchange dataset.

In addition, we compute correlation coefficient matrices between the historical variables and future variables of these two datasets. We also randomly select 100 sub-series of length 96 from the series, and calculate their correlation with the sub-series of the subsequent 96 time steps. The average of the 100 correlation matrices is also calculated. For correlation coefficients greater than 0.8, we assign the corresponding value to 1, otherwise, we assign the value to 0. Figure \ref{fig:similar_different} shows that the correlation between historical and future variables in the Electricity dataset is strong, indicating that one variable's historical series may be useful for predicting another variable's future series. However, for the Exchange dataset, the entire matrix is black. Even if we change the threshold of the correlation coefficient to 0.3, the matrix is still black, suggesting weak correlations between historical variables and future variables. This, we believe, is also the reason why it is difficult to achieve particularly accurate forecasting results in the Exchange dataset.

\section{Visualization of the attention matrices for Client}

\begin{figure}[!htbp]
% \vspace{-5pt}
  \includegraphics[scale=0.28]{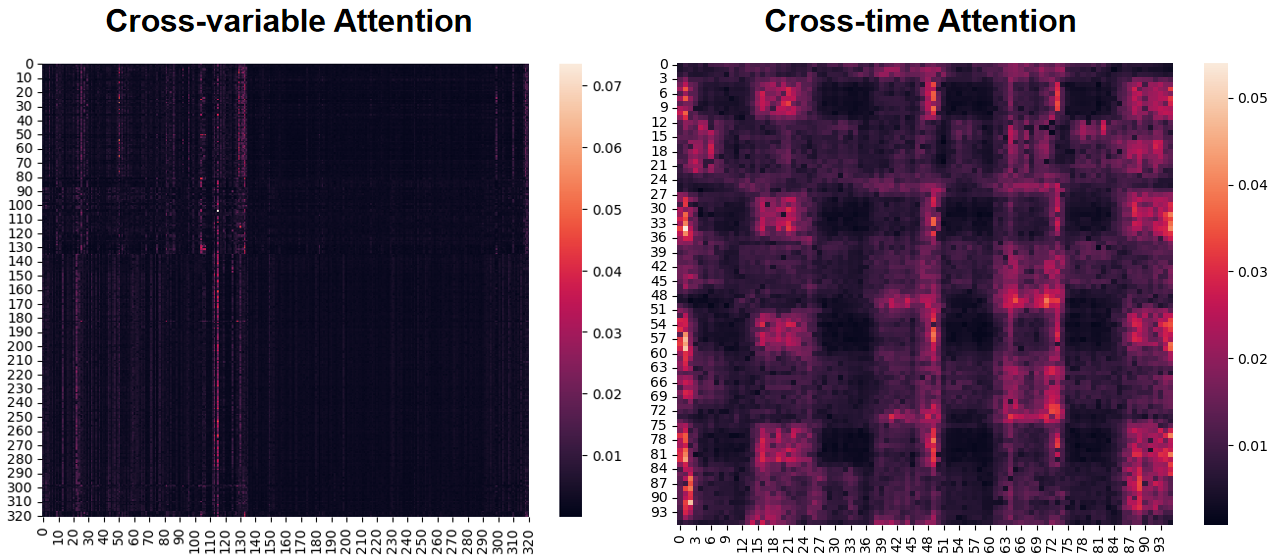}
  \centering
% \vspace{-5pt}
  \caption{Visualization of the attention matrices. The matrix on the left is Client's, and the matrix on the right is of cross-time based Transformer's.} %descripe in detail
  \label{fig:compare_att}
  % \vspace{-10pt}
\end{figure}

When compared with the Transformer model based on cross-time, Client's attention matrix, as depicted in \ref{fig:compare_att}, appears significantly sparser. This observation is justifiable since the predictions made by the model primarily depend on the information extracted from the variable series, and the inter-variable dependencies only assist in the predictive process. Consequently, the attention matrix of Client holds a relatively lesser amount of information than that of the cross-time Transformer, and thus, it is more sparse.

\section{Prediction Showcases}
\begin{figure}[!htbp]
\vspace{-150pt}
  \includegraphics[scale=0.38]{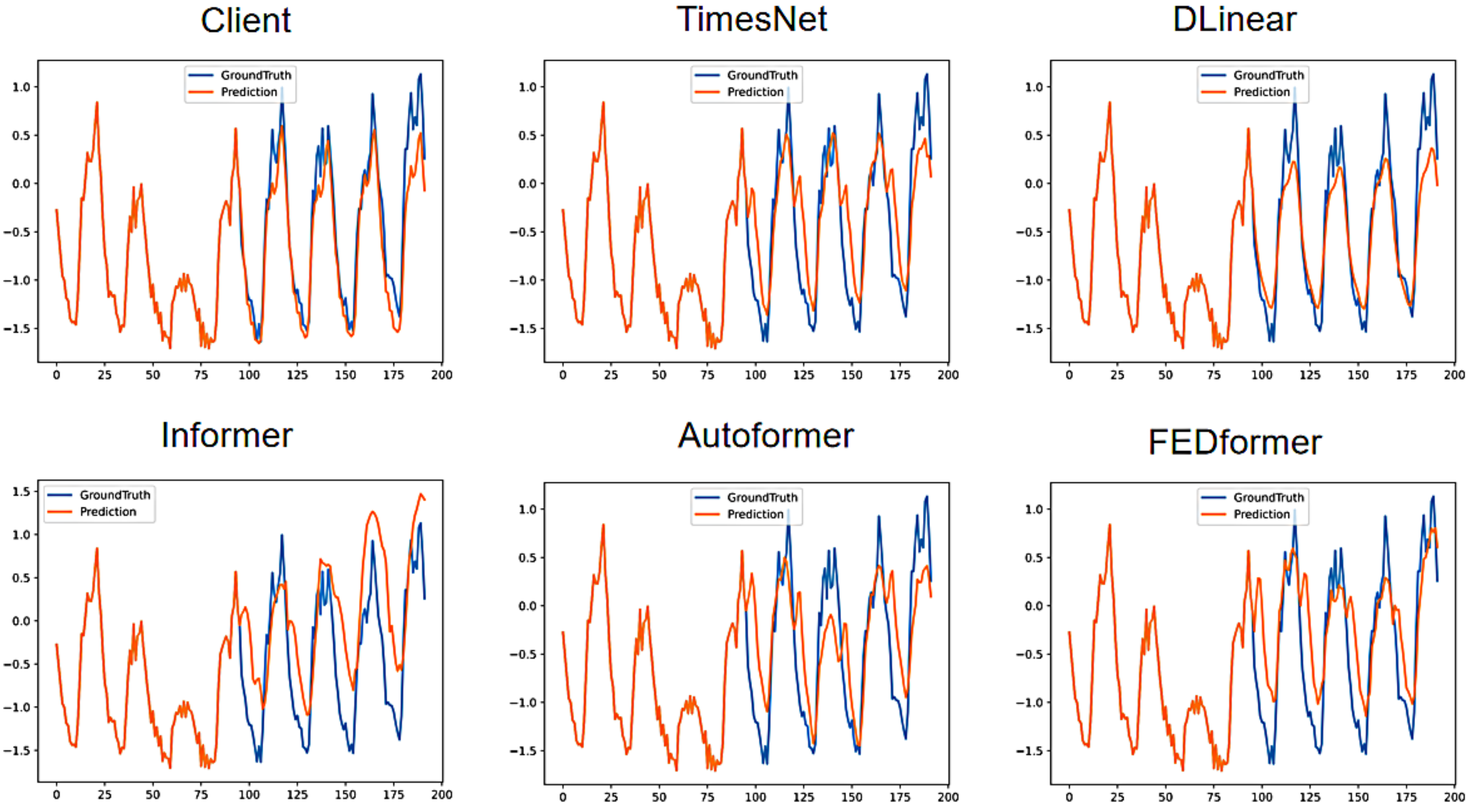}
  \centering
  \caption{Predictions comparison of different models of Electricity. Client outperforms the other models significantly, particularly regarding details.}
  \label{fig:compare_model}
\end{figure}

\begin{figure}[!htbp]
\vspace{-330pt}
  \includegraphics[scale=0.2]{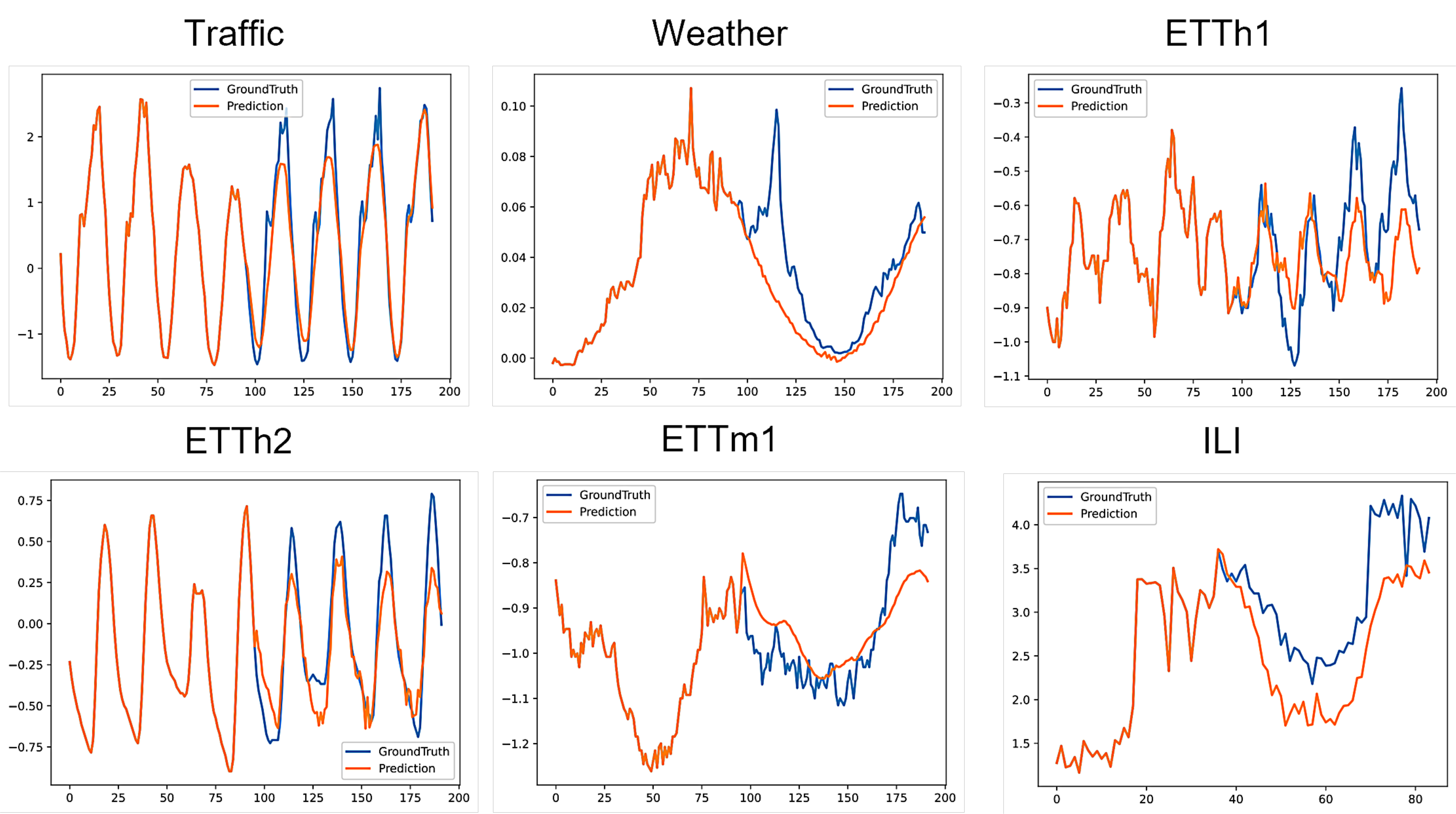}
  \centering
  \caption{The prediction performance of Client on different datasets. Client exhibits commendable results from the Traffic and ETTh2 datasets. However, there remains considerable scope for refining its performance on other datasets.}
  \label{fig:diff_data}
\end{figure}

We present the predictions of various models on identical samples in Figure \ref{fig:compare_model}, along with Client's forecasts on diverse datasets in Figure \ref{fig:diff_data}. Although Client demonstrates significant advantages over alternative models, there is still considerable scope for enhancing its predictive performance regarding certain datasets.


%%%%%%%%%%%%%%%%%%%%%%%%%%%%%%%%%%%%%%%%%%%%%%%%%%%%%%%%%%%%
\end{document}